\documentclass{bmvc2k}

\title{Attention Distillation: self-supervised vision transformer students need more guidance}
\addauthor{Kai Wang}{kwang@cvc.uab.es}{1}
\addauthor{Fei Yang (*corresponding author)}{fyang@cvc.uab.es}{1}
\addauthor{Joost van de Weijer}{joost@cvc.uab.es}{1}

\addinstitution{
 Computer Vision Center\\
 Universitat Autònoma de Barcelona\\
 Barcelona, Spain
}

\runninghead{Kai Wang et al.}{Attention Distillation}

\usepackage[utf8]{inputenc} 
\usepackage[T1]{fontenc}    
\usepackage{hyperref}       
\usepackage{url}            
\usepackage{booktabs}       
\usepackage{amsfonts}       
\usepackage{nicefrac}       

\usepackage[final]{pdfpages}
\usepackage{color}
\usepackage{xcolor}
\usepackage{graphicx}
\usepackage{pifont}
\usepackage{multirow}
\usepackage{xspace}
\usepackage{capt-of}
\usepackage{amsmath}
\usepackage{enumitem}
\usepackage{wrapfig,lipsum}
\usepackage{booktabs}       
\usepackage{microtype}      

\newcommand{\xmark}{\ding{55}\xspace}%
\usepackage{colortbl}
\definecolor{cyan}{cmyk}{.3,0,0,0}
\def\ourmethod{{AttnDistill}\xspace}

\newcommand{\tabincell}[2]{\begin{tabular}{@{}#1@{}}#2\end{tabular}}
\newcommand{\minisection}[1]{\vspace{0.02in} \noindent {\bf #1}\ }

\begin{document}

\maketitle

\begin{abstract}
    Self-supervised learning has been widely applied to train high-quality vision transformers (ViT). Unleashing their excellent performance on memory and compute constraint devices is therefore an important research topic.  However, how to distill knowledge from one self-supervised ViT to another has not yet been explored. Moreover, existing self-supervised knowledge distillation (SSKD) methods focus on ConvNet architectures and are suboptimal for ViT knowledge distillation. In this paper, we study knowledge distillation of self-supervised vision transformers (ViT-SSKD). We show that directly distilling information from the crucial attention mechanism from teacher to student can significantly narrow the performance gap between both. In experiments on ImageNet-Subset and ImageNet-1K, we  show that our method \ourmethod outperforms existing self-supervised knowledge distillation (SSKD) methods and achieves state-of-the-art \textit{k}-NN accuracy compared with self-supervised learning (SSL) methods learning from scratch (with the ViT-S model). We are also the first to apply the tiny ViT-T model for self-supervised learning. Moreover, \ourmethod is independent of self-supervised learning algorithms, and it can be adapted to ViT based SSL methods to improve performance in future research.
\end{abstract}

\section{Introduction}
Vision transformers~\cite{dosovitskiy2020image} have been widely applied in computer vision tasks, including image classification~\cite{touvron2021going,wu2021cvt,yuan2021tokens}, object recognition~\cite{carion2020end,dai2021up,fang2021you,sun2021rethinking,zhu2020deformable} and semantic segmentation~\cite{cheng2021per,strudel2021segmenter,xie2021segformer,zheng2021rethinking}. ViTs contain a self-attention mechanism~\cite{vaswani2017attention} that allows for information exchange between distant patches and consequently leads to a more holistic understanding of image content. Another important aspect of transformers is that they are often pretrained in a self-supervised manner, followed by a finetuning stage to adapt to the downstream task~\cite{devlin2018bert,li2022efficient}.  ViTs suffer from high memory requirements and substandard optimizability~\cite{chen2021mobile,dai2021coatnet,graham2021levit,mehta2021mobilevit,wang2021pyramid,yu2022unified}, making them unsuitable for applications on memory or computation constraint devices. Consequently, methods that can reduce the memory footprint while maintaining the performance of ViTs are in demand.

One transfer learning technique is knowledge distillation~\cite{hinton2015distilling}. 
Initial works focussed on knowledge transfer for networks trained in a supervised manner~\cite{chen2021distilling,tung2019similarity,yuan2020revisiting,ye2021lifelong}. Recently, the theory was extended to distill knowledge of self-supervised feature representations generated by large networks~\cite{fang2020seed,navaneet2022simreg,noroozi2018boosting}. Since these networks do not output a conditional probability over a label set, but rather a feature representation, alternative distillation techniques needed to be developed~\cite{abbasi2020compress,fang2020seed}. With the advent of transformers, supervised knowledge distillation for transformers has recently been investigated~\cite{jia2021efficient,liu2022meta,deit}. However, methods that can transfer self-supervised ViTs to smaller variants have not yet been explored. 

Therefore, we explore knowledge distillation of self-supervised ViTs. We find that existing theory designed to transfer knowledge of ConvNets trained in a self-supervised manner results in a significant performance gap between teacher and student.
To address this problem, we explore attention distillation that focuses on transferring the information present in the self-attention mechanism. Rather than just communicating the teacher's conclusion which is the focus of most traditional knowledge distillation methods, attention distillation provides more guidance to the student network by identifying the important regions for understanding the image content. The potential of attention distillation has been explored for ConvNet~\cite{gou2021knowledge,zagoruyko2016paying}, however, since \textit{for these networks attention is not explicitly computed, additional computation and attention definition are needed.} 
Since the attention mechanism is an integral and crucial part of transformers and \textit{no additional} computation is required, we argue that attention distillation is a \textit{natural extension} of the existing distillation theory for transformer networks. 

In this paper, we focus on \textit{self-supervised knowledge distillation of self-supervised vision transformers} (ViT-SSKD). First, we propose to use a \textit{projector alignment (PA)} module to align the class tokens from teacher and student models. 
Second, we propose \textit{attention guidance (AG)} with the Kullback–Leibler divergence to guide the student to obtain similar attention maps as the teacher model to further enhance the distillation. 
With these two modules, we can obtain state-of-the-art performance compared with self-supervised algorithms. 
Furthermore, we are
the \textit{first} to successfully train a small ViT-T model based on self-supervised learning (SSL) with knowledge distillation. 
More importantly, there might be more complex and outperforming SSL pretrained models in the future. In that case our method can be \textit{applied directly} to obtain a smaller model while keeping competitive performance. Our main contributions  are:

\begin{itemize}
    \item We are the first to study the important ViT-SSKD problem allowing to transfer knowledge to small transformers in a self-supervised fashion.
    \item We propose an attention distillation loss for improved guidance of the student during knowledge distillation. Our method, \ourmethod, significantly reduces the gap between teacher and student models. 
    \item We are the first to train a self-supervised ViT-T model. It obtains a performance almost (-0.3\%) at par with the supervised ViT-T model.
\end{itemize}


\section{Related work}
\minisection{Self-supervised learning.}
SSL~\cite{chen2020simple,chen2021exploring,doersch2015unsupervised,gidaris2018unsupervised,grill2020bootstrap,he2021masked,he2020momentum,van2018representation,wu2018unsupervised} automatically derives a supervisory signal for the training of high-quality feature representations, preventing the need of large labeled datasets. The common paradigm here is to pretrain on ImageNet~\cite{russakovsky2015imagenet} and then evaluated on downstream tasks, on which it has reached excellent performance, closing the gap with supervised methods. 
Recent popular SSL methods can be divided into two streams. \textit{Contrastive learning}~\cite{hadsell2006dimensionality,chen2020simple,caron2021emerging,he2020momentum} is the most popular stream.
Another stream of representation learning, named \textit{masked image encoding}~\cite{bao2021beit,dong2021peco,he2021masked,zhou2021ibot}, learns representations from corrupted images. 
\textit{In this paper, we study knowledge distillation for SSL based on both technical streams.}
From the aspect of backbone architectures, the previous methods are all based on ConvNet~\cite{chen2020simple,caron2020unsupervised,grill2020bootstrap}. Recently, with the appearance of ViT,  these are also applied for SSL (DINO~\cite{caron2021emerging}, MoCo~\cite{chen2021empirical}, etc). Compared with ConvNet, the attention-based ViTs suffer less from an image-specific inductive bias and have a larger potential when training on large scale datasets. 
\textit{In this paper we focus on the original ViT design, but the method could also be generalized to Swin Transformer~\cite{liu2021swin} based SSL methods~\cite{li2022efficient,xie2021self}.}

\minisection{Knowledge distillation for self-supervised models.}
Most knowledge distillation~\cite{hinton2015distilling} techniques are proposed under the supervised learning scenario~\cite{romero2014fitnets,tung2019similarity,peng2019correlation,zagoruyko2016paying,park2019relational,ahn2019variational,yuan2020revisiting,chen2021distilling,tian2019contrastive,zhao2022decoupled}.
Under the \textit{SSL} settings, CC~\cite{noroozi2018boosting} exploit pseudo labels from clustering teacher embeddings as distillation signals. 
Then SEED~\cite{fang2020seed} and CompRess~\cite{abbasi2020compress} maintain memory banks to store a huge number of samples to calculate instance-level similarity score distributions for aligning the teacher and student models. 
SimReg~\cite{navaneet2022simreg} has similar projector architecture as our method, where they use the projector to align the teacher and student features. However, 
in some cases when the student ViT architectures become quite different from the teacher model, only projector regression is not sufficient to  transfer knowledge from the teacher to the student model. 
Reg~\cite{yu2019learning} is specified for metric learning, which could also be applied to self-supervised representation distillation.
Recently, KDEP~\cite{he2022knowledge} propose the  power temperature scaling to distill representation from a supervised teacher model. 

Except for these examples in computer vision, there are several distillation attempts in NLP~\cite{wang2020minilm,sun2020mobilebert,jiao2019tinybert,fang2021compressing}. However, these methods are limited to the case that teacher-student models share similar architectures.  Also, Pelosin et al.~\cite{pelosin2022towards} apply attention distillation between similar transformer architectures for continual learning. Our proposal is a more generalizable framework and allows for attention distillation between different ViT architectures.

\section{Methodology}
\subsection{Preliminaries}
\begin{figure}[t]
\centering
\includegraphics[width=0.99\textwidth]{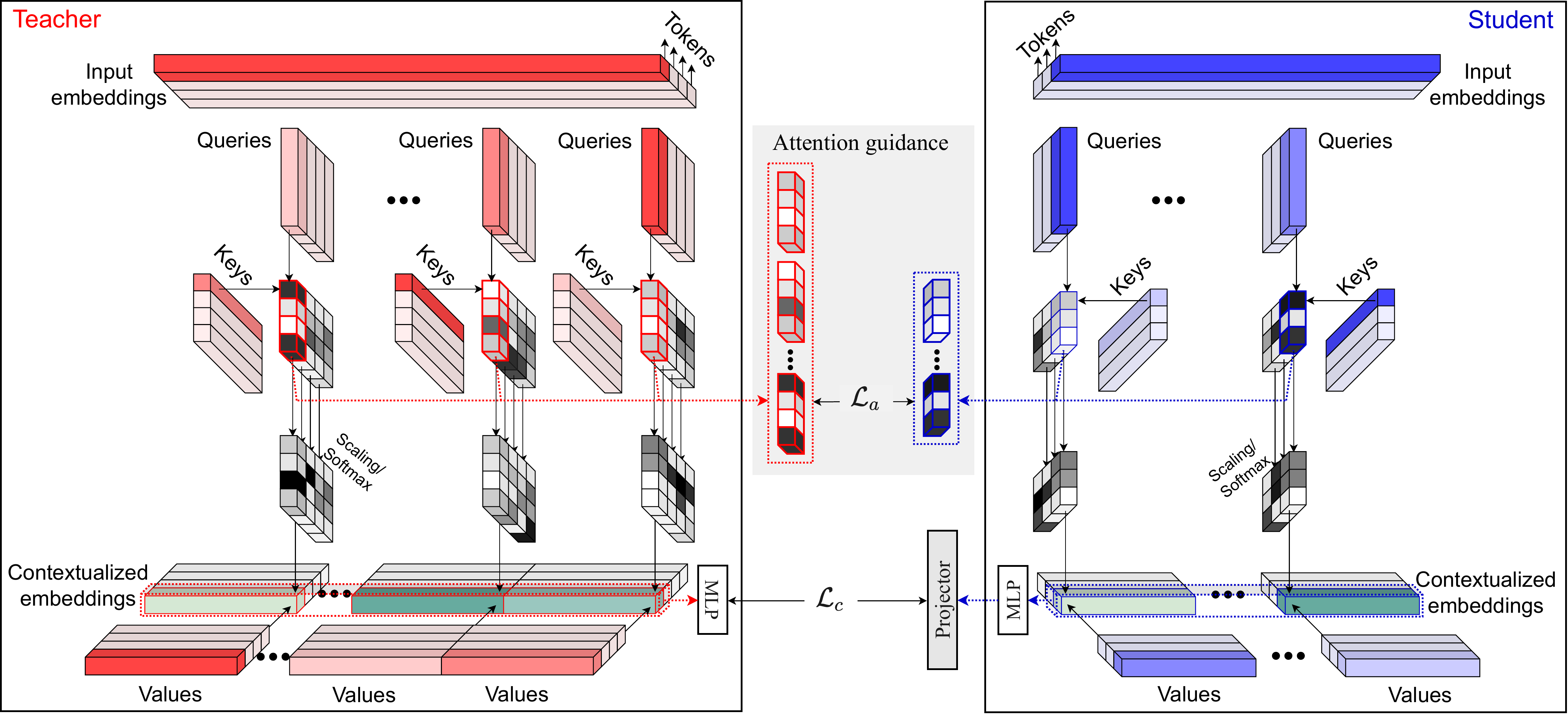}
\caption{\ourmethod for ViT-SSKD on the last block of the ViT. It is composed of the projector alignment loss $\mathcal{L}_c$ and attention guidance loss $\mathcal{L}_a$. The class tokens are taken from the last layer of the teacher and student ViT.
We only consider the attention vectors that are formed by the interaction of the class token query with all keys for distillation. 
}
\label{fig:vit_kd_2}
\end{figure}

\minisection{Vision Transformers Architecture.}
Here we consider the ViT proposed in~\cite{dosovitskiy2020image} but the theory is general and can be extended to other transformer architectures. The ViT consists of a patch embedding part, where the  transformer encoder  is a stack of $L$ 
multi-head self-attention blocks (MH-SAB). In each MH-SAB, there are two parts: a multi-head self-attention module (MSA) and a fully connected feedforward module (MLP). Each self-attention module has $H$ \textit{heads}. We will further use $d$ as the output dimension for each head and $N$ as the number of patches. Also considering the class token, we can denote the output of the $l$th MH-SAB as  $\mathbf{z}^l \in \mathbb{R}^{(N+1) \times (dH)}, l \in [1,L]$. $\mathbf{z}^0$ is the encoded patch embedding of image $\mathbf{x}$ (i.e., from $\mathbf{z}_{1}^0$ to $\mathbf{z}_{N}^0$) and the initial class token (i.e., $\mathbf{z}_{0}^0$). For the $h$th head in the self-attention module, learnable parameters $W_Q^{l,h},W_K^{l,h},W_V^{l,h}$ implemented as FC layers, map one slice of the input tokens $\mathbf{z}^{l-1,h}$ into the queries, keys and values ($Q^{l,h}, K^{l,h},V^{l,h} \in \mathbb{R}^{(N+1) \times (d)}$). We obtain the attention map with Eq.~\ref{eq:msa_1}, where $\mathbb{A}^{l,h} \in \mathbb{R}^{(N+1) \times (N+1)}$, and the output of this head is obtained by Eq.~\ref{eq:msa_2}.
Combining the multiple head outputs, we obtain the final output of this multi-head self-attention layer (see Eq.~\ref{eq:msa_3}).
Finally, with the MLP and layer normalization (LN), the output tokens produced by the $l$th MH-SAB are given by Eq.~\ref{eq:msa_4}. We have shown the process of the last MH-SAB from both the teacher and student ViT separately in Fig.~\ref{fig:vit_kd_2} ({the design is inspired from~\cite{vit_fig}.}) which includes the key components in our approach.
\begin{align}
  \label{eq:msa_1}
  \mathbb{A}^{l,h} & = \text{Softmax}(Q^{l,h} \cdot (K^{l,h})^T/\sqrt{d}) \\
  \label{eq:msa_2}
  \mathbf{y}^{l,h} &= \mathbb{A}^{l,h} \cdot V^{l,h}\\
  \label{eq:msa_3}
  \mathbf{y}^{l} &= concat(\mathbf{y}^{l,1}, \mathbf{y}^{l,2}, ..., \mathbf{y}^{l,H})\\
  \label{eq:msa_4}
  \mathbf{z}^{l} &= \text{MLP}\left(\text{LN}\left(\mathbf{y}^{l} + \mathbf{z}^{l-1}\right)\right) + \mathbf{z}^{l-1}.
\end{align}

\subsection{\ourmethod: Attention distillation}
An outline of \ourmethod is given in Fig.~\ref{fig:vit_kd_2}. It can be divided into two parts: projector alignment (PA) and attention guidance (AG). 
For our distillation, we do not need to generate multiple views, since we focus on distilling the knowledge of the teacher. So other than contrastive-based SSL methods~\cite{caron2021emerging,zhou2021ibot,zhou2022mugs}, in this phase\textit{ we do not rely on multi-crop augmentations}. Thus, we have less computation costs and \textit{the effective epoch is equal to the training epoch.}



\minisection{Projector alignment (PA).} Suppose we have a teacher-student pair, each with a ViT architecture named ${V}_{t}$ and $V_{s}$. In self-supervised knowledge distillation, our aim is to distill knowledge from the teacher to the student model in a SSL way while maintaining its transferability. 
In most cases, we expect a smaller student model compared to the teacher. \textit{The parameter size is highly dependent on the feature dimension in the ViT. Thus, ${V}_{t}$ and $V_{s}$ typically have different feature dimensions.} Therefore, we introduce a linear mapping projector $\mathcal{P}$ to map the student to the teacher feature space for alignment. And since in ViT, \textit{the class token embedding} $\mathbf{E}^{c}=\mathbf{z}^L_0$ is the most representative embedding for a classification decision, in \ourmethod, we only map the class token from the last layer for aligning the teacher and student model with a MSE loss to \textit{communicate} with the final output from the teacher:
\begin{equation}
  \label{eq:cls_mse}
  \mathcal{L}_{c} = ||  \mathbf{E}^{c}_{t} - \mathcal{P}(\mathbf{E}^{c}_{s}) ||_2
\end{equation}
where `s/t' subscripts represent `student/teacher'.
In our ablation study, we also explored the influence of aligning the image patch embeddings $\mathbf{z}^L_i, i \in [1,N]$ with the linear mapping $\mathcal{P}$. 

\begin{figure}[t]
\centering
\includegraphics[width=0.95\textwidth]{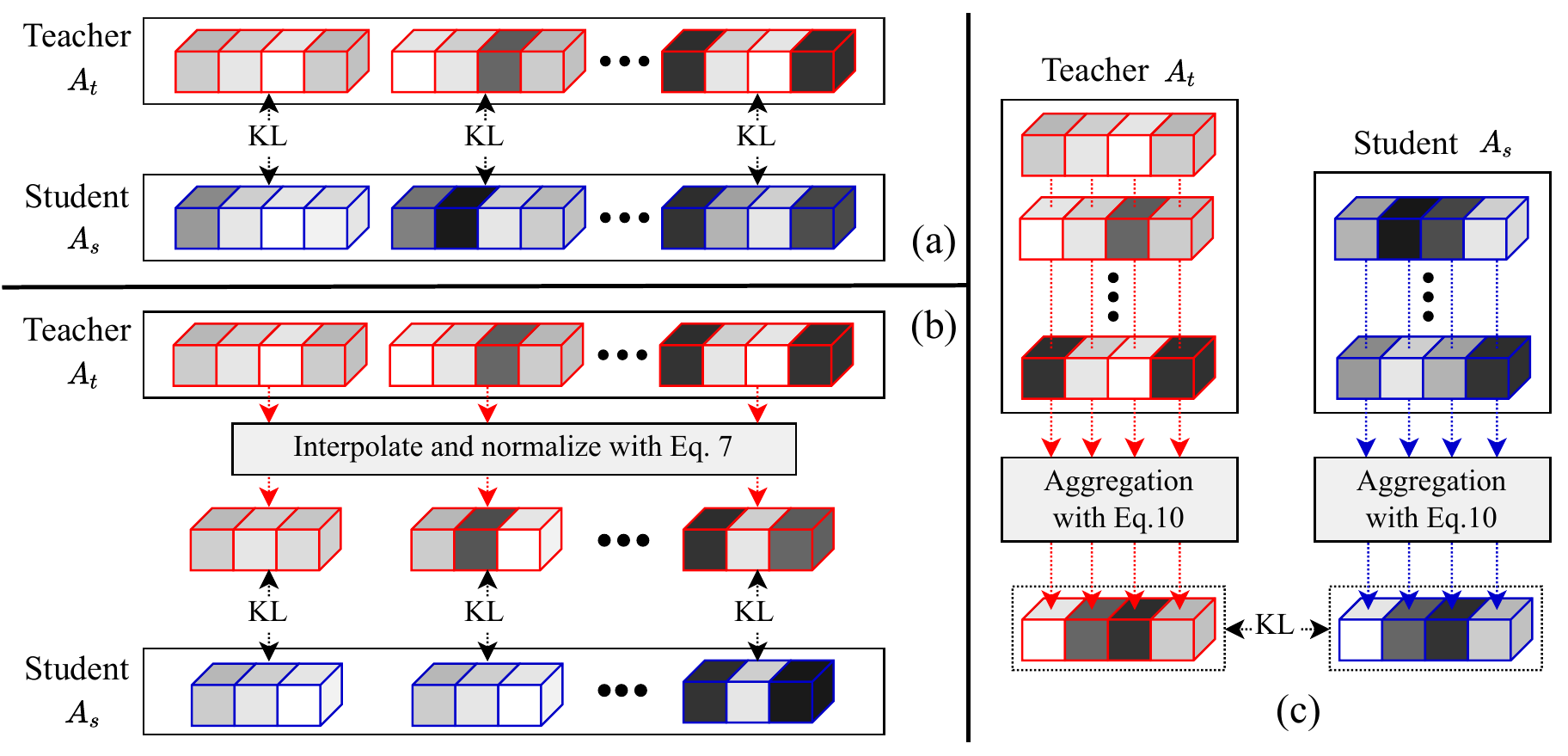}
\caption{ Attention guidance between varying transformer architectures.}
\label{fig:attn_kd_3}
\end{figure}

\minisection{Attention guidance (AG).}However, aligning the class tokens can only tell the student model about "what" is in the image. More guidance from the teacher explaining "why" it reached this conclusion could be helpful. This extra guidance can be extracted from the multi-head self-attention (MSA) mechanism in ViTs. The MSA module pays attention to the decisive and informative parts in the image. Actually, we can observe that distillation with other methods  will lead to \emph{attention drift} where the student attention differs from the teacher (see the attention maps in Fig.~\ref{fig:single_train_aggr}.)

For a ViT, each $\mathbb{A}^{l,h}$ given by layer $l$ head $h$ is an attention map from all tokens to all tokens. And since $\mathbb{A}^{L,h}_{0,j}, j \in [0,N]$ ({`0' represents the first row of the attention map $A^{L,h}$.}) contains the attention probabilities for the class token, it represents the importance of each token  for the classification prediction of the image. By denoting $A=\mathbb{A}^{L}_{0}, A^h=\mathbb{A}^{L,h}_{0}, a^h_j=\mathbb{A}^{L,h}_{0,j}$, we propose to apply Kullback–Leibler divergence ($\mathcal{KL}$) to make the student model pay attention to the same regions as the teacher model by aligning $A_{s}$ and $A_{t}$. 
We also study the performance with attentions from all layers in our ablation study section.
To address knowledge transfer between ViTs with different designs, several cases need to be addressed.  
Here, we categorize them into four cases that might occur and discuss the solution below. The illustration of these variations are provided in Fig.~\ref{fig:attn_kd_3}.\\
\begin{enumerate}[label=(\alph*), leftmargin=*]
    \item    The teacher and student models have the same number of heads $H=H_{t}=H_{s}$ and patches $N=N_{t}=N_{s}$ (see Fig.2 (a)). 
    This is the simplest case, where we align according to:
    \begin{equation}
      \label{eq:case1}
      \mathcal{L}_{a} = \sum_{h \in [1,H]} \mathcal{KL}( A_{t}^h || A_{s}^h )
    \end{equation}
    
    \item  The teacher and student models have the same number of heads $H$ but a different number of patches $N_{t}$ and $N_{s}$ (see Fig.2 (b)). In this case, we propose to interpolate ($\mathbf{IP}$, {by default we apply bicubic function~\cite{keys1981cubic} as a smoother interpolation}) the teacher model attention map $(a^h_j)_{t}, j \in [1,N_{t}]$ into $(a^h_j)_{t}^{'}, j \in [1,N_{s}]$ ($N_t=w \times h, N_s=w' \times h'$), and then normalize ($\mathbf{NR}$) them into $1-(a^h_0)_{t}$ by scaling up to have the attentions sum to 1 as in Eq.~\ref{eq:interpolate}.  Then the \textit{attention guidance} loss is given by Eq.~\ref{eq:interpolate_adloss}.
    \begin{align}
      \label{eq:interpolate}
      (a^h_j)_{t}^{'} & = \mathbf{NR}_{1-(a^h_0)_{t}}(\mathbf{IP}((a^h_j)_{t})) \\
      \label{eq:interpolate_adloss}
      \mathcal{L}_{a} &= \sum_{h \in [1,H]} \mathcal{KL}( (A^h)^{'}_{t} || A_{s}^h )
    \end{align}
    
    \item  The teacher and student models have the same number of patches $N$ but a different number of heads $H$ (see Fig.2 (c)). Here we merge the attentions from all heads for distillation. We have considered several aggregation functions, including mean, maximum and soft-maximum. 
    We found that using the $log$ summation to aggregate the attention probabilities for both teacher and student models (see Eq.~\ref{eq:aggr1} and Eq.~\ref{eq:aggr2}) leads to slightly superior results compared to max-based fusion.
    Then the \textit{attention guidance} loss $\mathcal{L}_{a}$ is as Eq.~\ref{eq:aggr_adloss}.
    \begin{align}
        \label{eq:aggr1}
        a_j=&  \frac{1}{T} \cdot \sum_{h \in [1,H]} log(a^h_j) = \frac{1}{T} \cdot  log(\prod_{h \in [1,H]} a^h_j) \\
        \label{eq:aggr2}
        A = &\mathtt{Softmax} ([a_0,a_1,...,a_{N}]) \\
      \label{eq:aggr_adloss}
       \mathcal{L}_{a} = &\mathcal{KL}( A_{t} || A_{s} )
    \end{align}
    This aggregation could effectively highlight the maximum probabilities from the attention maps of the $H$ heads, as can be seen from our ablation study.\\
    
    \item   The teacher and student models have a different number of heads $H$ and patches $N$. This case is a combination of the above two, thus we apply interpolation and aggregation sequentially and then apply distillation.
\end{enumerate}

Finally, the self-supervised loss to update the student model $V_{s}$ is:
    \begin{equation}
      \label{eq:final_loss}
      \mathcal{L} = \mathcal{L}_{c} + \lambda \cdot \mathcal{L}_{a}
    \end{equation}
\section{Experiments}
\subsection{Pre-Training setup}
\minisection{Datasets.} In our experiments, ImageNet-Subset~\cite{russakovsky2015imagenet} is used for ablation study and to compare with other self-supervised knowledge distillation methods. 
This dataset contains 100 classes and $\approx$130k images in high resolution (resized to 224$\times$224) ~\cite{rebuffi2017icarl}. For comparison with  SSL methods, we employ the ImageNet-1K dataset~\cite{russakovsky2015imagenet}. 

\minisection{Architecture.} For the Teacher-Student pairs, we focus on knowledge distillation from a larger ViT teacher model to a smaller ViT student model. Due to the high computation demands of ViT, we select Teacher-Student pairs as below:
\begin{itemize}
    \item On ImageNet-1K we select the following three pairs (Teacher \textrightarrow Student): (a) Mugs(ViT-S/16) \textrightarrow ViT-T/16; (b)  Mugs(ViT-B/16) \textrightarrow ViT-S/16; (c)  DINO(ViT-S/8) \textrightarrow ViT-S/16;
    \item On ImageNet-Subset, we fix the  teacher model as MAE(ViT-S/16) with 12-Layer, 6-Head, 16-Patch and vary the design of the student model in the ablation study.
\end{itemize}
Moreover, since MAE/MoCo-v3 and DINO/iBOT/Mugs are following different position embedding strategies (fixed vs. learnable), our teacher-student pairing setups can show the effectiveness of \ourmethod for these two kind of position encodings.

\minisection{Implementation details.} We train our \ourmethod with the AdamW~\cite{loshchilov2017decoupled} optimizer. The learning rate is linearly ramped up during the first 40 epochs to the base learning rate $lr=(1.5e-4) \times \text{batchsize}/256$. After the warming up epochs, we decay it with a cosine schedule till 800 epochs (except the distillation from Mugs(ViT-S/16) to ViT-T/16, where we train for 500 epochs because of performance saturation). 
By default, we set $T=10.0$ and $\lambda=0.1$, the projector $\mathcal{P}$ is a 4-layer linear mapping. For the evaluations of the student model,  we found it is optimal to perform with the features before the $\mathcal{P}$.
For the experiments on ImageNet-Subset, we fix the teacher model as a ViT-S/16 pre-trained with MAE~\cite{he2021masked} method with 3200 epochs.
More details are in our supplementary materials.

\begin{table}[!t]
\begin{minipage}[c]{.63\linewidth}
\centering
\setlength{\tabcolsep}{1.2mm}
{
\scalebox{0.58}
{
\begin{tabular}{cllccccc}
\toprule
Teacher model & Method & Student Arch. & Par.(M)  & Train Epo. & Effect Epo. & $k$-NN & LP. \\
\midrule

\textcolor{gray!80}{\xmark} & \textcolor{gray!80}{Supervised} & \textcolor{gray!80}{ViT-T/16} & \textcolor{gray!80}{5.7} & \textcolor{gray!80}{-} & \textcolor{gray!80}{-}  & \textcolor{gray!80}{72.2} & \textcolor{gray!80}{72.2}\\

SwAV (RN-50) & CRD &  RN-18 & 11 & 240 & 240    & 44.7 & 58.2 \\

SwAV (RN-50) & CC &  RN-18 & 11 & 100 & 100 & 51.0 & 60.8 \\
SwAV (RN-50) & Reg &  RN-18 & 11 & 100 & 100    & 47.6 & 60.6 \\
SwAV (RN-50) & CompRess-2q &  RN-18 & 11 & 130  & 130   & 53.7 & 62.4 \\
SwAV (RN-50) & CompRess-1q &  RN-18 & 11 & 130  & 130   & 56.0 & 65.6 \\
SwAV (RN-50) & SimReg &  RN-18 & 11 & 130 & 130 & 59.3 & 65.8 \\

SwAV (RN-50$\times$2) & SEED &  RN-18 & 11 & 200  & 200   & 55.3 & 63.0 \\
SwAV (RN-50$\times$2) & SEED &  EffNet-B1 & 7.8  & 200 & 200   & 60.3 & 68.0 \\
SwAV (RN-50$\times$2) & SEED &  EffNet-B0 & 5.3  & 200 & 200   & 57.4 & 67.6 \\
SwAV (RN-50$\times$2) & SEED &  MbNet-v3 & 5.5  & 200 & 200   & 55.9 & 68.2 \\

\rowcolor{cyan!50}
Mugs (ViT-S/16) & \ourmethod & ViT-T/16 & 5.7  & 500  & 500 & \textbf{71.4} & \textbf{71.9} \\

\midrule
\textcolor{gray!80}{\xmark} &\textcolor{gray!80}{Supervised} & \textcolor{gray!80}{ViT-S/16} & \textcolor{gray!80}{22} & \textcolor{gray!80}{-} & \textcolor{gray!80}{-} &\textcolor{gray!80}{79.8} & \textcolor{gray!80}{79.8}\\
{\xmark} &SimCLR  & ViT-S/16 & 22  & 300 & 600  & - & 69    \\ 
{\xmark} &BYOL    & ViT-S/16 & 22  & 300 & 600  & - & 71    \\ 

{\xmark} &MoCo v3 & ViT-S/16 & 22  & 600  &1200 & - & 73.4  \\ 
{\xmark} &SwAV    & ViT-S/16 & 22  & 800 & 2400 & 66.3 & 73.5  \\ 
{\xmark} &DINO    & ViT-S/16 & 22  & 800 & 3200 & 74.5 & 77.0    \\ 

{\xmark} &iBOT    & ViT-S/16 & 22  & 800 & 3200 & 75.2 & 77.9  \\ 
{\xmark} &MUGS    & ViT-S/16 & 22  & 800 & 3200 & 75.6 & \textbf{78.9}  \\
SwAV (RN-50$\times$2) & SEED &  RN-34 & 21 & 200 & 200    & 58.2 & 65.7 \\
SwAV (RN-50$\times$2)  & SEED &  RN-50 & 24 & 200 & 200   & 59.0 & 74.3 \\

SimCLR (RN-50$\times$4)  & CompRess-1q &  RN-50 & 24 & 130 & 130   & 63.3 & 71.9 \\
SimCLR (RN-50$\times$4)  & CompRess-2q &  RN-50 & 24 & 130 & 130   & 63.0 & 71.0 \\

SimCLR (RN-50$\times$4)  & CC &  RN-50 & 24 & 100 & 100   & 55.6 & 68.9 \\

SimCLR (RN-50$\times$4)  & SimReg &  RN-50 & 24 & 130 & 130 & 60.3 & 74.2 \\

\rowcolor{cyan!50}
Mugs (ViT-B/16) &\ourmethod & ViT-S/16 & 22   & 800 & 800 & {76.8} & 78.6 \\

\rowcolor{cyan!50}
DINO (ViT-S/8) &\ourmethod & ViT-S/16 & 22   & 800 & 800 & \textbf{77.4} & 78.8 \\

\midrule
\multicolumn{8}{l}{\textit{\textbf{Teacher Models statistics}}} \\
Mugs (ViT-S/16) & -    & - & 22  & 800 & 3200 & 75.6 & {78.9}  \\
DINO (ViT-S/8) & -    & - & 22  & 800 & 3200 & 78.3 & {79.7}  \\

Mugs (ViT-B/16) & -    & - & 85  & 400 & 1600 & 78.0 & 80.6  \\
SwAV (RN-50) & - & - & 24  & 800  & 2400 & 64.8 & 75.6 \\
SwAV (RN-50$\times$2)  & - & - & 94 & 800  & 2400 & - & 77.3 \\
SimCLR (RN-50$\times$4) & - & - & 375 & 1000& 2000 & 64.5 & 75.6 \\

\bottomrule
\end{tabular}
}
}
\caption{ Comparison with state-of-the-art SSL methods with $k$-NN and linear probing (LP.) on \textbf{ImageNet-1K}. "Effect Epo." is the effective pretraining epochs computed by multiplying number of views processed by the models following iBOT~\cite{zhou2021ibot}.
}
\label{tab:linear}
\end{minipage}
\hspace{1mm}
\begin{minipage}{.35\linewidth}
\centering
\includegraphics[width=0.9\textwidth]{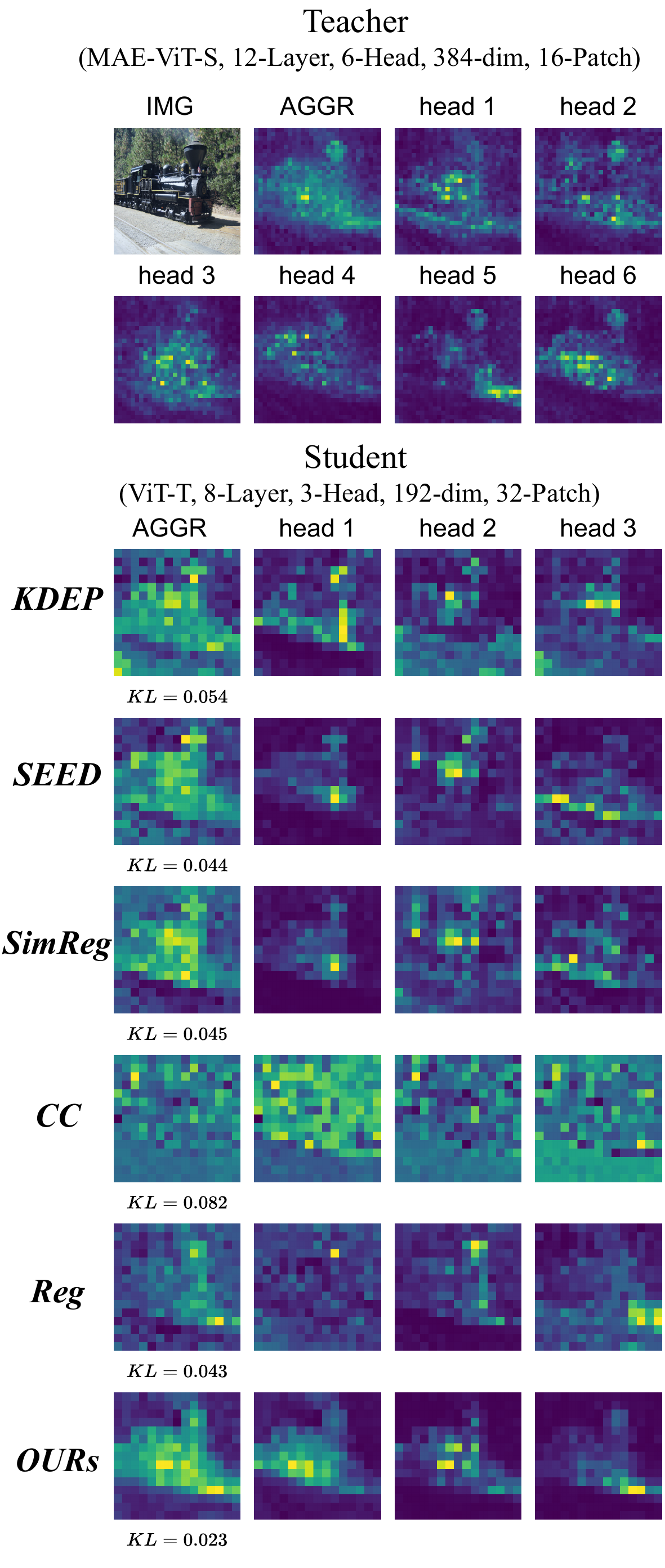}
\captionof{figure}{For the teacher model, we show the original image, the aggregated attention map (AGGR) and the attention maps for each head. The KL distances to the teacher AGGR are shown.}
\label{fig:single_train_aggr}
\end{minipage}
\end{table}

\subsection{Comparison with state-of-the-art}
\minisection{Comparison with SSKD methods on ViT.} As shown in Table~\ref{tab:sskd}, we compare several methods to distill a  MAE(ViT-S/16) teacher to a ViT-T student on \textit{ImageNet-Subset}. Next to the common ViT-T architecture (with 12-layers, 6-heads, 16-patches), we also consider a harder variant (with 8-layers, 3-heads, 32-patches) as the student model. \ourmethod clearly outperforms them all in both cases. The margin is larger for the harder setup. This also indicates the importance of the attention distillation guidance. 
A comparison of the attention maps from the 8-layer 3-head 32-patch student case is shown in Fig.~\ref{fig:single_train_aggr}.

\minisection{Comparison with SSL methods with linear probing, $k$-NN and finetuning accuracies.} For linear probing and $k$-NN evaluation on \textit{ImageNet-1K}, we follow the commonly used setting in DINO~\cite{caron2021emerging} and iBOT~\cite{zhou2021ibot}. The comparison is shown in Table~\ref{tab:linear}. We draw the following conclusions:
\begin{itemize}[leftmargin=*]
    \item Based on ViT-T/16 distilled from Mugs(ViT-S/16), our method \ourmethod gets state-of-the-art $k$-NN and Linear probing performance compared with previous knowledge distillation methods based on ConvNet. \ourmethod(ViT-T/16) is with only 5.7M parameters but outperforms the previous methods by a large margin and gets quite close to the supervised ViT-T/16 learning from scratch. In this case, we only train the student model for 500 epochs since we have observed marginal improvement on linear probing after that.
    \item Based on ViT-S/16 distilled from DINO(ViT-S/8), our method \ourmethod gets state-of-the-art in $k$-NN and the second in linear probing. 
    This distillation decreases the ViT computational demand since there are 75\% less patches in ViT-S/16 than ViT-S/8. Then, based on ViT-S/16 distilled from Mugs(ViT-B/16), \ourmethod gets the second in $k$-NN and the third in linear probing evaluations. In this case, the model size is decreased by 75\%.
\end{itemize}

\begin{figure*}[t]
\begin{center}
\begin{tabular}{ccc}
\includegraphics[width=0.31\textwidth]{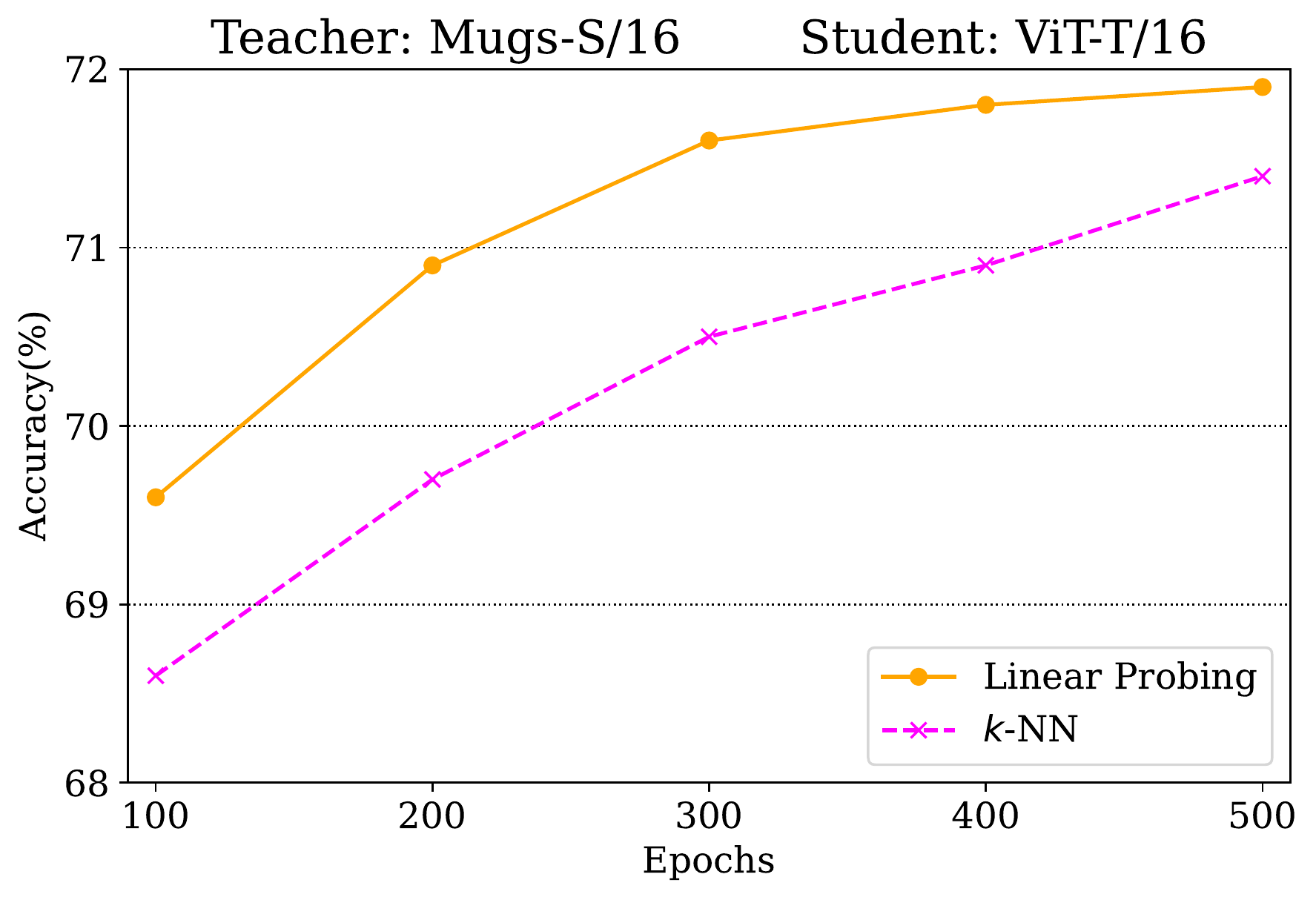} &
\includegraphics[width=0.31\textwidth]{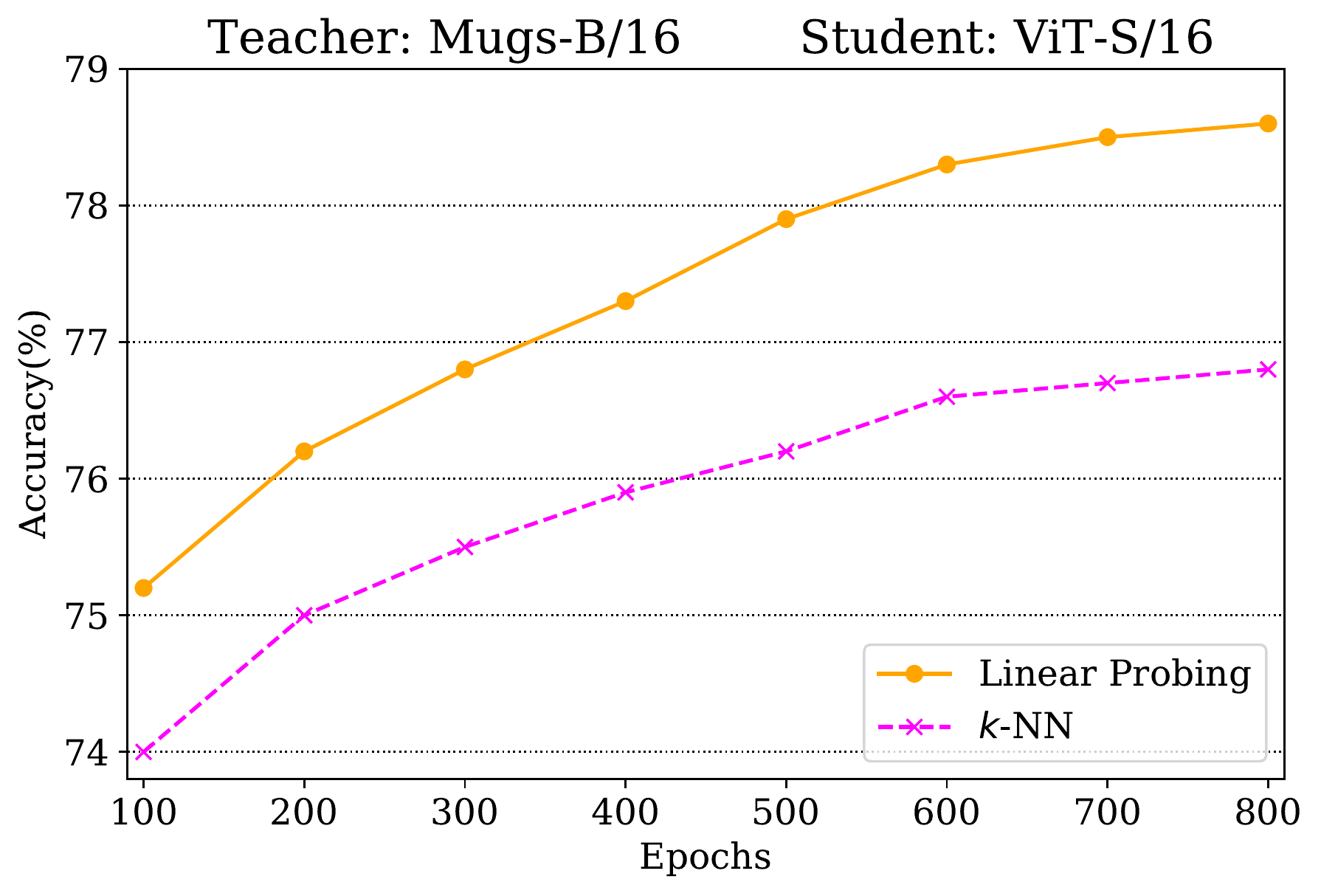} &
\includegraphics[width=0.31\textwidth]{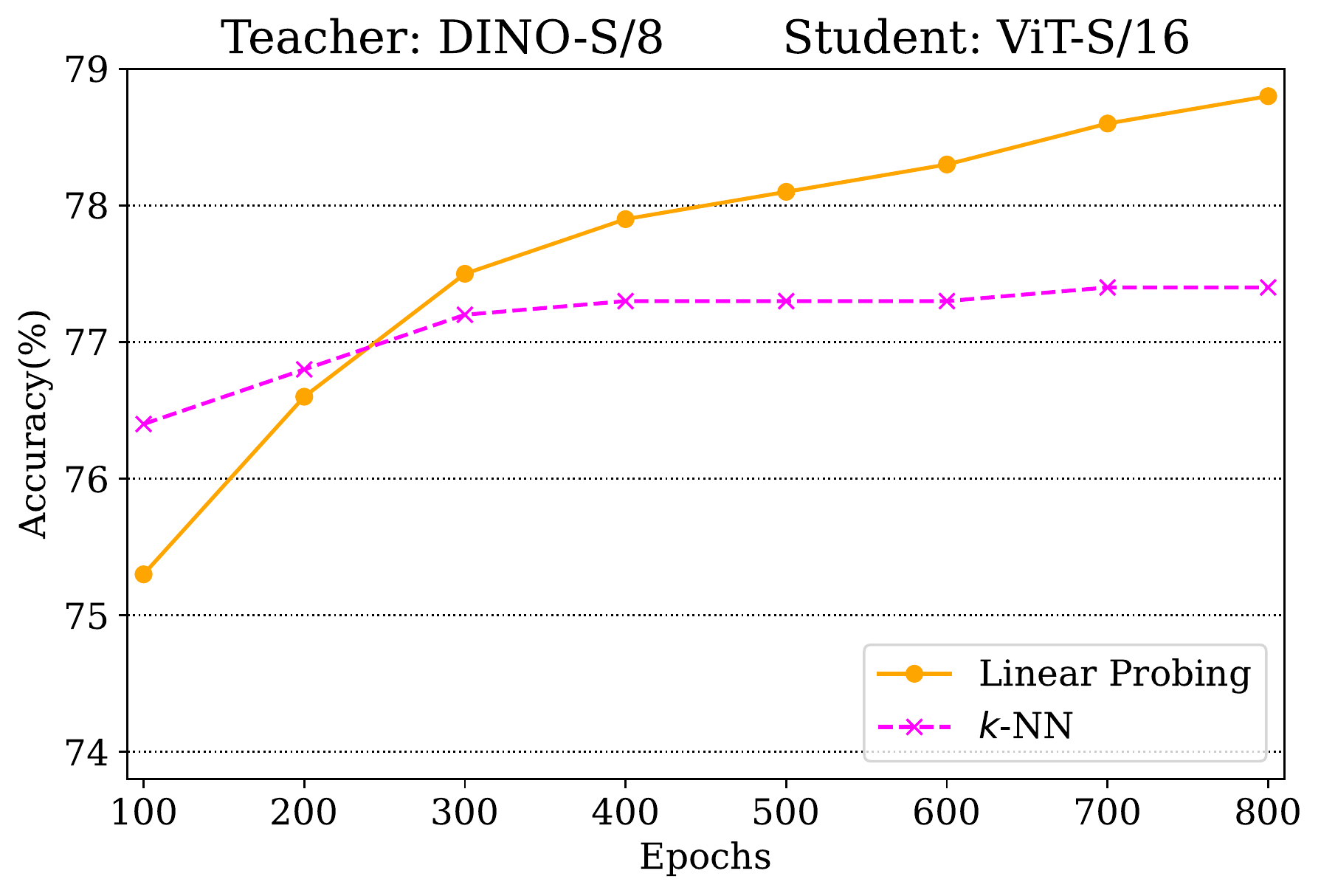}
\end{tabular}
\end{center}
\caption{More results of \ourmethod with different training epochs on \textbf{ImageNet-1K}.
}
\label{fig:acc_epoch}
\end{figure*}

Moreover, we could further observe the advantage of ViT in $k$-NN evaluations, which means the extracted features from self-supervised pretrained ViT are more beneficial without learning an extra classifier as in linear probing evaluations. 
The accuracy curves during training are shown in Fig.~\ref{fig:acc_epoch}.

For the finetuning comparison on ImageNet-1K shown in Table~\ref{tab:finetuning}, we compared with existing methods  working on ViT-T and ViT-S. \ourmethod (ViT/T) and \ourmethod (ViT/S) are distilled from Mugs (ViT-B) and Mugs (ViT-S) respectively. In both cases, \ourmethod works better than supervised learning methods and just marginally worse than the state-of-the-art with ViT-S. 
In conclusion, whereas \ourmethod is state-of-the-art for $k$-NN evaluation (Table~\ref{tab:linear}), this is not the case when evaluating by means of finetuning.

\minisection{Comparison with SSL methods on downstream tasks.} For semi-supervised learning, results with SSL methods based on ViT-S/16 are shown in Table~\ref{tab:semi_cls}. In this setting, first, models are trained self-supervised on all \textit{ImageNet-1K} data. Next labels for a small fraction of data (1\% or 10\%) are used to perform fine-tuning, linear probing or \textit{k}-NN classification. 
Under all three settings with only 1\% of the data, we can observe a considerable advantage of \ourmethod  with a 3.9\%/2.6\%/5.8\% improvement compared with Mugs(ViT-S/16). With 10\% data, the improvement is less but still notable as 0.4\%/2.7\%/3.1\%.

We also evaluate \ourmethod (ViT-T) for transfer learning. We compare with supervised ViT models, since there are no papers using ViT-T for SSL. Results are summarized in Table~\ref{tab:cifar_vit_tiny}) for several small datasets. \ourmethod gets a 2.3\% improvement compared with the previous best supervised learning method CCT-7/3x1~\cite{hassani2021escaping}. Next, we consider transfer learning of \ourmethod (ViT-S) in Table~\ref{tab:cifar_vit_small}. Here we compare with previous SSL methods, we are only marginally worse than the state-of-the-art and much better than the supervised distillation method DEiT~\cite{deit}.


\begin{table}[!tb]
\begin{minipage}[c]{.31\linewidth}
\centering
\setlength{\tabcolsep}{3.5mm}{
\scalebox{0.45}{
\begin{tabular}{lcccc}
\toprule
 Method & Top-1 & Top-5  & Top-1  & Top-5  \\

\midrule
\tabincell{l}{Student\\ViT-T Arch.} & \multicolumn{2}{c}{{12-L, 6-H, 16-P}} & \multicolumn{2}{c}{{8-L, 3-H, 32-P}}  \\ 
\midrule

{\tabincell{l}{Teacher\\MAE-S/16}}
& {79.4 }& {93.6}   & {79.4 }& {93.6} \\ 
\midrule
SEED~\cite{fang2020seed}    & 71.7 & 90.7 & 66.8 & 88.4 \\ 
CompRess~\cite{abbasi2020compress}  & 71.9  & 90.8  & 67.2 & 88.7 \\ 

CC~\cite{noroozi2018boosting}      & 63.8 & 87.6 & 47.1 & 74.9 \\
KDEP~\cite{he2022knowledge} & 66.7 & 87.8  & 57.8 &  82.4 \\

Reg~\cite{yu2019learning}  & 76.2 & 92.5 & 69.6 & 89.0 \\
SimReg~\cite{navaneet2022simreg}  & 77.8 & 93.4  & 68.6 & 88.8 \\

\midrule
\rowcolor{cyan!50}
\ourmethod  & \textbf{79.3} & \textbf{94.1}  & \textbf{73.8} & \textbf{91.7}  \\ 
\bottomrule
\end{tabular}}
}
\caption{Compare with SSKD methods on ImageNet Subset with top-1/top-5 (LP.)}
\label{tab:sskd}
\end{minipage}
\hspace{1mm}
\begin{minipage}[c]{.31\linewidth}
\centering
\setlength{\tabcolsep}{1.2mm}
{
\scalebox{0.52}
{
\begin{tabular}{lccc}
\toprule
 Method  & Arch.(M) & Effect Epo. & FT Acc. \\
\midrule
\multicolumn{3}{l}{\textit{Supervised learning}} \\
 - & ViT-T/16 & - & 72.2 \\
 - & ViT-S/16 & - & 79.8 \\
 DeiT & ViT-S/16 & - & 81.3 \\
 DeiT-III~\cite{Touvron2022DeiTIR} & ViT-S/16 & - & 81.4 \\
 Manifold~\cite{jia2021efficient} & ViT-S/16 & - & 81.5 \\
 MKD~\cite{liu2022meta} & ViT-S/16 & - & 82.1 \\
 \midrule
 \multicolumn{3}{l}{\textit{Self-Supervised learning}} \\
 MoCo v3 & ViT-S/16 & 600 & 81.4 \\
 DINO & ViT-S/16 & 3200 & 82.0 \\
 iBOT & ViT-S/16 & 3200 & 82.3 \\
 Mugs & ViT-S/16 & 3200 & \textbf{82.6} \\
 \rowcolor{cyan!50}
 \ourmethod  & ViT-T/16 & 500 & 72.9 \\
 \rowcolor{cyan!50}
 \ourmethod  & ViT-S/16 & 800 & 81.6 \\

\bottomrule
\end{tabular}}
}
\caption{Finetuning comparison on ImageNet1K.
}
\label{tab:finetuning}
\end{minipage}%
\hspace{1mm}
\begin{minipage}[c]{0.31\linewidth}
\centering
\setlength{\tabcolsep}{1.2mm}
{
\scalebox{0.47}
{
\begin{tabular}{lc cc cc  cc}
\toprule
\multirow{2}{*}{Method} & \multirow{2}{*}{Arch}  & \multicolumn{2}{c}{{FT}} & \multicolumn{2}{c}{{LP}} & \multicolumn{2}{c}{{\textit{k}-NN}}\\

 &  & 1\% & 10\%  & 1\% & 10\% & 1\% & 10\%\\
\midrule

SimCLR & RN50  & 57.9 & 68.1 & - & - & - & -\\
BYOL & RN50  & 53.2 & 68.8 & - & - & - & -\\
SwAV & RN50  & 53.9 & 70.2  & - & -& - & -\\
SimCLR+SD & RN50 & 60.0 & 70.5  & - & -& - & -\\

DINO & ViT-S/16   & 60.3 & 74.3  & 59.1 & 70.3 & 61.2 & 69.0 \\ 

iBOT & ViT-S/16  & 61.9 & 75.1  & 61.5 & 71.8 & 62.5 & 70.1 \\ 

Mugs & ViT-S/16  & 66.8 & 76.8 & 64.1 & 72.2 & 63.6 & 70.6\\ 


\rowcolor{cyan!50}
\ourmethod   & ViT-S/16  & \textbf{70.7} & {\textbf{77.2}}  & \textbf{66.7} & \textbf{74.9} & \textbf{69.4} & \textbf{73.7} \\ 
\bottomrule
\end{tabular}}
}
\caption{Semi-supervised learning on ImageNet1K. \ourmethod (ViT-S) is distilled from the teacher model Mugs (ViT-B/16) for 800 epochs.
}
\label{tab:semi_cls}
\end{minipage}%
\end{table}

\begin{table}[!t]
\begin{minipage}[c]{.45\linewidth}
\centering
\setlength{\tabcolsep}{1.2mm}
{
\scalebox{0.7}
{
\begin{tabular}{lccc}
\toprule
Method  & Par.(M) & CIFAR100 & CIFAR10 \\
\midrule
\multicolumn{3}{l}{\textit{Supervised learning}} \\

SL-CaiT~\cite{lee2021vision}  & 9.2  & 80.3 & 95.8  \\ 

SL-T2T~\cite{lee2021vision}    & 7.1  & 77.4 & 95.6  \\ 

SL-Swin~\cite{lee2021vision}   & 10.2 & 80.0 & 95.9  \\ 

CVT-7/4~\cite{hassani2021escaping}  & 3.7 & 73.0 & 92.4  \\ 

CCT-7/3x1~\cite{hassani2021escaping}  & 3.8 & 82.7 & 98.0    \\
\midrule
\multicolumn{3}{l}{\textit{Self-Supervised + Transfer learning}} \\

\rowcolor{cyan!50}
\ourmethod (ViT-T)  & 5.7 & \textbf{85.0} & \textbf{98.1}  \\ 
\bottomrule
\end{tabular}}
}
\caption{Transfer learning comparison on CIFAR10/CIFAR100. \ourmethod (ViT-T) is distilled from Mugs(ViT-S/16).}
\label{tab:cifar_vit_tiny}
\end{minipage}%
\hspace{1mm}
\begin{minipage}[c]{.5\linewidth}
\centering
\setlength{\tabcolsep}{1.2mm}
{
\scalebox{0.64}
{
\begin{tabular}{lccccc}
\toprule
Method  & Par.(M) & CIFAR100 & CIFAR10 & Flowers & Cars \\
\midrule
\multicolumn{6}{l}{\textit{Supervised + Transfer learning}} \\
- & 22 & 89.5& 99.0 & 98.2 & 92.1 \\
DeiT-III~\cite{Touvron2022DeiTIR} & 22 & 90.6 & 98.9 & 96.4 & 89.9 \\
\midrule
\multicolumn{6}{l}{\textit{Self-Supervised  + Transfer learning}} \\
BEiT & 22 & 87.4 & 98.6 & 96.4 & 92.1 \\ 
DINO & 22 & 90.5 & 99  & 98.5 & 93.0  \\ 
iBOT & 22 & 90.7 & 99.1 & 98.6 & \textbf{94.0 }\\ 
Mugs & 22 & \textbf{91.8} & \textbf{99.2} & \textbf{98.8} & 93.9 \\ 
\rowcolor{cyan!50}
\ourmethod (ViT-S)  & 22 & 91.6 & 99.1 & 98.6 & 93.8 \\ 
\bottomrule
\end{tabular}}
}
\caption{Compared with SSL methods on four small datasets. \ourmethod (ViT-S) is distilled from the teacher Mugs (ViT-B/16).
}
\label{tab:cifar_vit_small}
\end{minipage}
\end{table}

\subsection{Ablation study}

To prove the generalizability of \ourmethod, we perform an ablation study on \textit{ImageNet-Subset} with a fixed MAE(ViT-S/16) teacher and vary the architecture of the student model.  Extended ablation studies are  in the supplementary. In Fig.~\ref{fig:ablation_study}, our ablation study contains three parts:

\begin{enumerate}[label=(\alph*),leftmargin=*]
    \item \minisection{The architecture of ViT (in Fig.~\ref{fig:ablation_study}-(a))}: To verify the effectiveness of \ourmethod for various architectures of ViT, we modify the number of heads, patch sizes and the number of block layers. In all cases, \ourmethod significantly improves the PA baseline and closes the gap with the teacher performance. Especially, for the smaller student architectures and those with fewer tokens, attention distillation is shown to be crucial leading to improvements of over 5\%. 
    
    \item     \minisection{The various aggregation functions (in Fig.~\ref{fig:ablation_study}-(b))}: Here we fix the design of the student model and vary the strategy to compute the attention guidance loss. To verify the superiority of the used log summation in Eq.~\ref{eq:aggr1}, we replace it with \textit{MEAN/MIN/MAX} strategies to aggregate attention maps from different heads. However, they are all suboptimal.
   
    \item \minisection{Alternative self-supervised losses (in Fig.~\ref{fig:ablation_study}-(c))}: A recent work for distillation of self-supervised representations of ConvNet is CompRess~\cite{abbasi2020compress}.
    Here, we apply the knowledge distillation loss from CompRess~\cite{abbasi2020compress} to our PA module, we can clearly observe that this KD loss is an obstacle in ViT distillation since when combined with our PA module it leads to a performance drop. Except distilling the attention maps from the last layer, we also experiment the distillation over attention maps from all layers. This is 0.6\% lower than \ourmethod based on only the last layer. Finally, we also align the patch tokens with our PA module. This is 2.4\% worse than without the patch token alignment, thus aligning patch tokens is not necessary.
\end{enumerate}

\begin{figure*}[t]
\begin{center}
\begin{tabular}{ccc}
\includegraphics[width=0.31\textwidth]{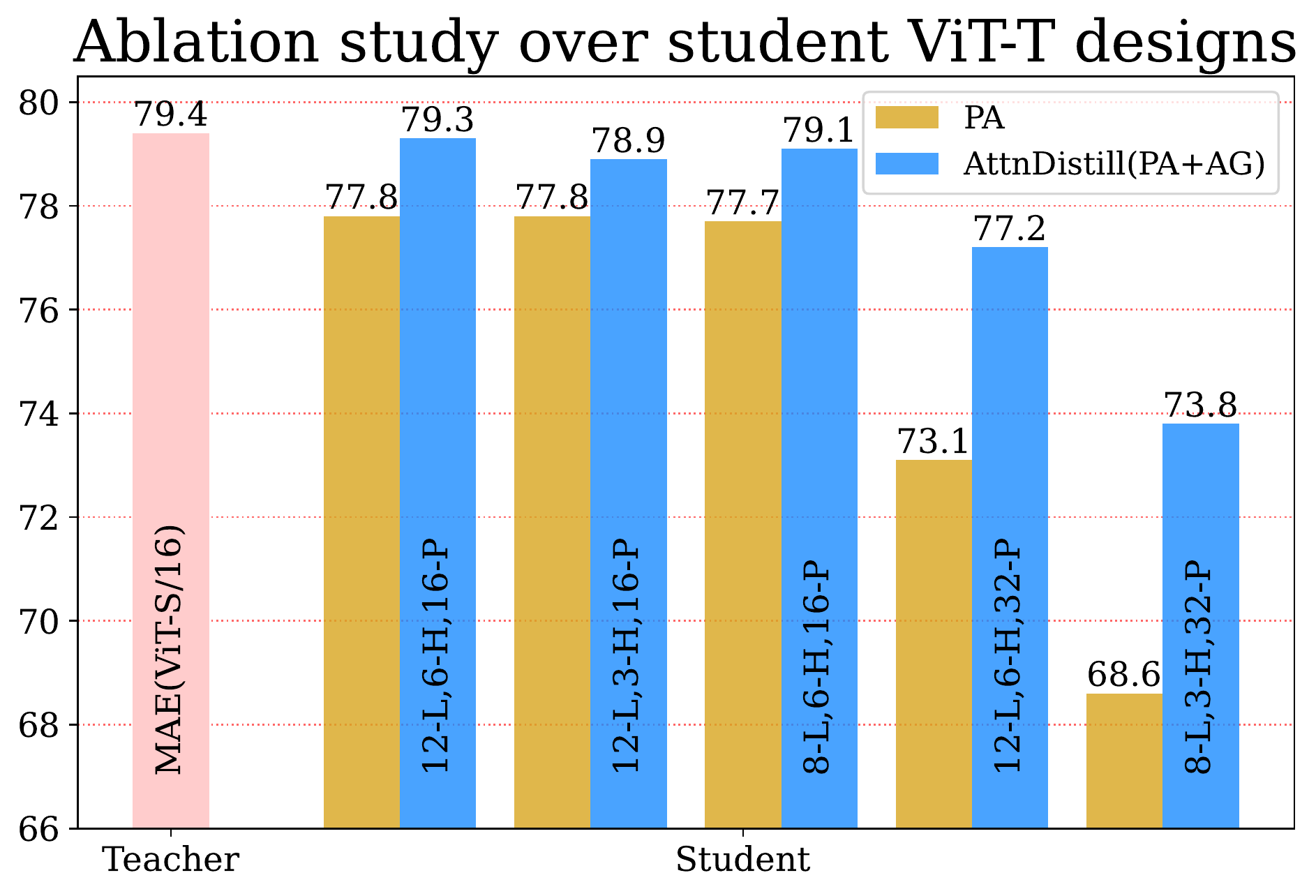} &
\includegraphics[width=0.31\textwidth]{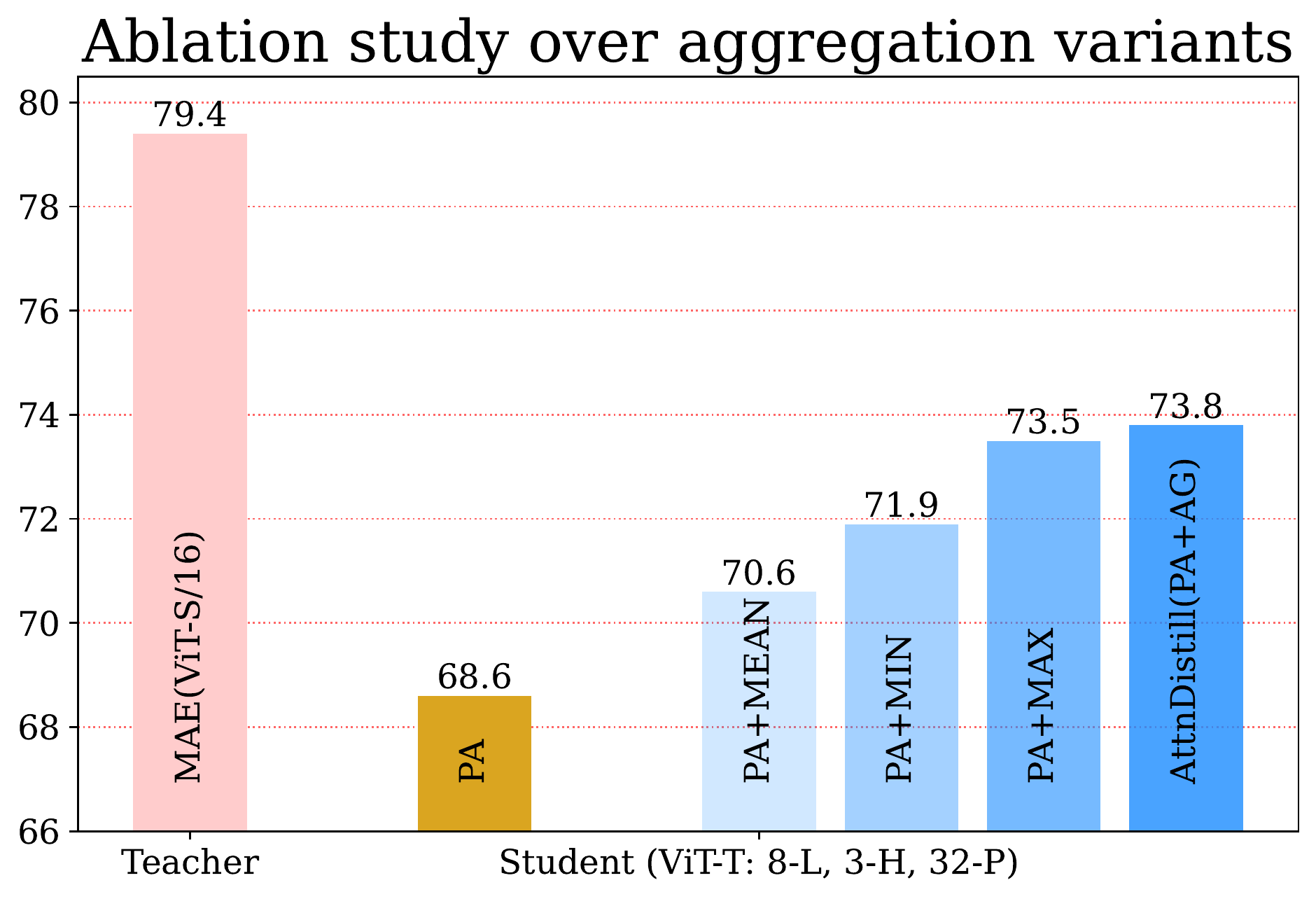} &
\includegraphics[width=0.31\textwidth]{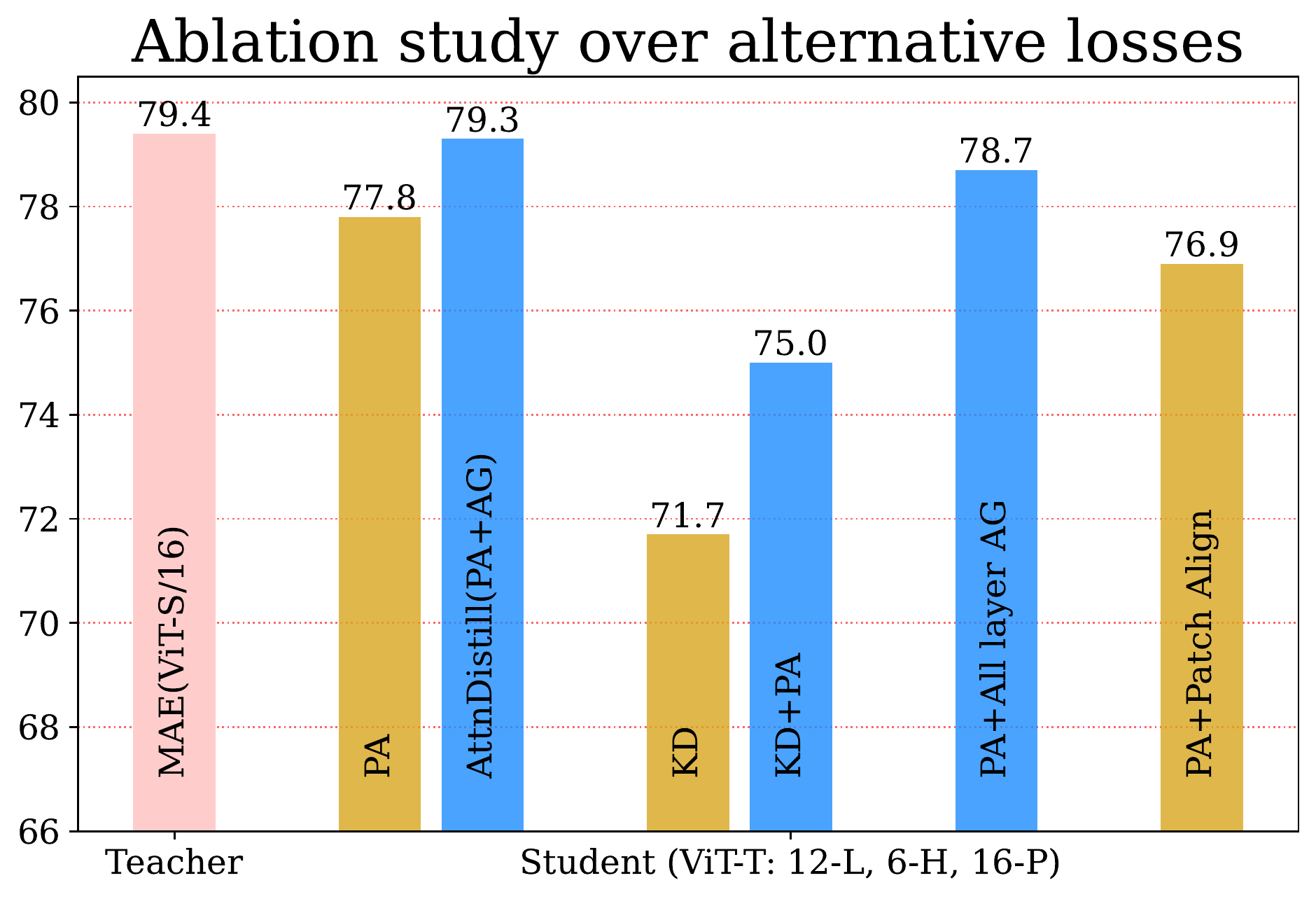}\\
(a) ViT architectures & (b) aggregation functions & (c) alternative losses
\end{tabular}
\end{center}
\caption{Ablation study for ViT architectures, aggregation functions and alternative losses.
}
\label{fig:ablation_study}
\end{figure*}

\section{Conclusion}
In this paper, we explored the ViT-based self-supervised knowledge distillation problem. Observing that the previous SSKD methods focussed on ConvNet do not work well on ViT, we proposed \textit{\ourmethod} to distill the knowledge from a pretrained teacher model to its student model. The experiments clearly show that  \ourmethod outperforms other SSKD methods. Furthermore, our distilled ViT-S gets state-of-the-art in k-NN accuracy and is second in linear probing compared with SSL methods. Also, our method \ourmethod is especially advantageous in semi-supervised learning evaluation and competitive in transfer learning evaluation. 
To prove the effectiveness of \ourmethod, we also implement various ablation studies on ImageNet-Subset. 
For future work, we are interested to explore \ourmethod for knowledge distillation between ConvNets and ViT.

\minisection{Limitations.}A drawback of the attention mechanism is that it is tailored for transformer usage and requires additional computation when applied to ConvNets (namely the computation of the attention maps). A further limitation is that the theory only applies to a single teacher-student pair. In case of multiple teacher models, further thought has to be given on how the multiple attention maps can be meaningfully communicated with the student.

\section*{Acknowledgement}
We acknowledge the support from Huawei Kirin
Solution, and the Spanish Government funded project PID2019-104174GB-I00/AEI/10.13039/501100011033.

\bibliography{longstrings,egbib}

\begin{thebibliography}{76}
\providecommand{\natexlab}[1]{#1}
\providecommand{\url}[1]{\texttt{#1}}
\expandafter\ifx\csname urlstyle\endcsname\relax
  \providecommand{\doi}[1]{doi: #1}\else
  \providecommand{\doi}{doi: \begingroup \urlstyle{rm}\Url}\fi

\bibitem[Abbasi~Koohpayegani et~al.(2020)Abbasi~Koohpayegani, Tejankar, and
  Pirsiavash]{abbasi2020compress}
Soroush Abbasi~Koohpayegani, Ajinkya Tejankar, and Hamed Pirsiavash.
\newblock Compress: Self-supervised learning by compressing representations.
\newblock \emph{Advances in Neural Information Processing Systems},
  33:\penalty0 12980--12992, 2020.

\bibitem[Ahn et~al.(2019)Ahn, Hu, Damianou, Lawrence, and
  Dai]{ahn2019variational}
Sungsoo Ahn, Shell~Xu Hu, Andreas Damianou, Neil~D Lawrence, and Zhenwen Dai.
\newblock Variational information distillation for knowledge transfer.
\newblock In \emph{Proceedings of the IEEE/CVF Conference on Computer Vision
  and Pattern Recognition}, pages 9163--9171, 2019.

\bibitem[Bao et~al.(2022)Bao, Dong, and Wei]{bao2021beit}
Hangbo Bao, Li~Dong, and Furu Wei.
\newblock Beit: Bert pre-training of image transformers.
\newblock \emph{International Conference on Learning Representations}, 2022.

\bibitem[Carion et~al.(2020)Carion, Massa, Synnaeve, Usunier, Kirillov, and
  Zagoruyko]{carion2020end}
Nicolas Carion, Francisco Massa, Gabriel Synnaeve, Nicolas Usunier, Alexander
  Kirillov, and Sergey Zagoruyko.
\newblock End-to-end object detection with transformers.
\newblock In \emph{European conference on computer vision}, pages 213--229.
  Springer, 2020.

\bibitem[Caron et~al.(2020)Caron, Misra, Mairal, Goyal, Bojanowski, and
  Joulin]{caron2020unsupervised}
Mathilde Caron, Ishan Misra, Julien Mairal, Priya Goyal, Piotr Bojanowski, and
  Armand Joulin.
\newblock Unsupervised learning of visual features by contrasting cluster
  assignments.
\newblock \emph{Advances in Neural Information Processing Systems},
  33:\penalty0 9912--9924, 2020.

\bibitem[Caron et~al.(2021)Caron, Touvron, Misra, J{\'e}gou, Mairal,
  Bojanowski, and Joulin]{caron2021emerging}
Mathilde Caron, Hugo Touvron, Ishan Misra, Herv{\'e} J{\'e}gou, Julien Mairal,
  Piotr Bojanowski, and Armand Joulin.
\newblock Emerging properties in self-supervised vision transformers.
\newblock In \emph{Proceedings of the IEEE/CVF International Conference on
  Computer Vision}, pages 9650--9660, 2021.

\bibitem[Chen et~al.(2021{\natexlab{a}})Chen, Liu, Zhao, and
  Jia]{chen2021distilling}
Pengguang Chen, Shu Liu, Hengshuang Zhao, and Jiaya Jia.
\newblock Distilling knowledge via knowledge review.
\newblock In \emph{Proceedings of the IEEE/CVF Conference on Computer Vision
  and Pattern Recognition}, pages 5008--5017, 2021{\natexlab{a}}.

\bibitem[Chen et~al.(2020)Chen, Kornblith, Norouzi, and Hinton]{chen2020simple}
Ting Chen, Simon Kornblith, Mohammad Norouzi, and Geoffrey Hinton.
\newblock A simple framework for contrastive learning of visual
  representations.
\newblock In \emph{International conference on machine learning}, pages
  1597--1607. PMLR, 2020.

\bibitem[Chen and He(2021)]{chen2021exploring}
Xinlei Chen and Kaiming He.
\newblock Exploring simple siamese representation learning.
\newblock In \emph{Proceedings of the IEEE/CVF Conference on Computer Vision
  and Pattern Recognition}, pages 15750--15758, 2021.

\bibitem[Chen et~al.(2021{\natexlab{b}})Chen, Xie, and He]{chen2021empirical}
Xinlei Chen, Saining Xie, and Kaiming He.
\newblock An empirical study of training self-supervised vision transformers.
\newblock In \emph{Proceedings of the IEEE/CVF International Conference on
  Computer Vision}, pages 9640--9649, 2021{\natexlab{b}}.

\bibitem[Chen et~al.(2022)Chen, Dai, Chen, Liu, Dong, Yuan, and
  Liu]{chen2021mobile}
Yinpeng Chen, Xiyang Dai, Dongdong Chen, Mengchen Liu, Xiaoyi Dong, Lu~Yuan,
  and Zicheng Liu.
\newblock Mobile-former: Bridging mobilenet and transformer.
\newblock \emph{Proceedings of the IEEE Conference on Computer Vision and
  Pattern Recognition}, 2022.

\bibitem[Cheng et~al.(2021)Cheng, Schwing, and Kirillov]{cheng2021per}
Bowen Cheng, Alex Schwing, and Alexander Kirillov.
\newblock Per-pixel classification is not all you need for semantic
  segmentation.
\newblock \emph{Advances in Neural Information Processing Systems}, 34, 2021.

\bibitem[Dai et~al.(2021{\natexlab{a}})Dai, Cai, Lin, and Chen]{dai2021up}
Zhigang Dai, Bolun Cai, Yugeng Lin, and Junying Chen.
\newblock Up-detr: Unsupervised pre-training for object detection with
  transformers.
\newblock In \emph{Proceedings of the IEEE/CVF Conference on Computer Vision
  and Pattern Recognition}, pages 1601--1610, 2021{\natexlab{a}}.

\bibitem[Dai et~al.(2021{\natexlab{b}})Dai, Liu, Le, and Tan]{dai2021coatnet}
Zihang Dai, Hanxiao Liu, Quoc~V Le, and Mingxing Tan.
\newblock Coatnet: Marrying convolution and attention for all data sizes.
\newblock \emph{Advances in Neural Information Processing Systems},
  34:\penalty0 3965--3977, 2021{\natexlab{b}}.

\bibitem[Devlin et~al.(2018)Devlin, Chang, Lee, and Toutanova]{devlin2018bert}
Jacob Devlin, Ming-Wei Chang, Kenton Lee, and Kristina Toutanova.
\newblock Bert: Pre-training of deep bidirectional transformers for language
  understanding.
\newblock \emph{arXiv preprint arXiv:1810.04805}, 2018.

\bibitem[Doersch et~al.(2015)Doersch, Gupta, and
  Efros]{doersch2015unsupervised}
Carl Doersch, Abhinav Gupta, and Alexei~A Efros.
\newblock Unsupervised visual representation learning by context prediction.
\newblock In \emph{Proceedings of the IEEE international conference on computer
  vision}, pages 1422--1430, 2015.

\bibitem[Dong et~al.(2021)Dong, Bao, Zhang, Chen, Zhang, Yuan, Chen, Wen, and
  Yu]{dong2021peco}
Xiaoyi Dong, Jianmin Bao, Ting Zhang, Dongdong Chen, Weiming Zhang, Lu~Yuan,
  Dong Chen, Fang Wen, and Nenghai Yu.
\newblock Peco: Perceptual codebook for bert pre-training of vision
  transformers.
\newblock \emph{arXiv preprint arXiv:2111.12710}, 2021.

\bibitem[Dosovitskiy et~al.(2020)Dosovitskiy, Beyer, Kolesnikov, Weissenborn,
  Zhai, Unterthiner, Dehghani, Minderer, Heigold, Gelly,
  et~al.]{dosovitskiy2020image}
Alexey Dosovitskiy, Lucas Beyer, Alexander Kolesnikov, Dirk Weissenborn,
  Xiaohua Zhai, Thomas Unterthiner, Mostafa Dehghani, Matthias Minderer, Georg
  Heigold, Sylvain Gelly, et~al.
\newblock An image is worth 16x16 words: Transformers for image recognition at
  scale.
\newblock \emph{International Conference on Learning Representations}, 2020.

\bibitem[Fang et~al.(2021{\natexlab{a}})Fang, Liao, Wang, Fang, Qi, Wu, Niu,
  and Liu]{fang2021you}
Yuxin Fang, Bencheng Liao, Xinggang Wang, Jiemin Fang, Jiyang Qi, Rui Wu,
  Jianwei Niu, and Wenyu Liu.
\newblock You only look at one sequence: Rethinking transformer in vision
  through object detection.
\newblock \emph{Advances in Neural Information Processing Systems}, 34,
  2021{\natexlab{a}}.

\bibitem[Fang et~al.(2021{\natexlab{b}})Fang, Wang, Hu, Wang, Yang, and
  Liu]{fang2021compressing}
Zhiyuan Fang, Jianfeng Wang, Xiaowei Hu, Lijuan Wang, Yezhou Yang, and Zicheng
  Liu.
\newblock Compressing visual-linguistic model via knowledge distillation.
\newblock In \emph{Proceedings of the IEEE/CVF International Conference on
  Computer Vision}, pages 1428--1438, 2021{\natexlab{b}}.

\bibitem[Fang et~al.(2021{\natexlab{c}})Fang, Wang, Wang, Zhang, Yang, and
  Liu]{fang2020seed}
Zhiyuan Fang, Jianfeng Wang, Lijuan Wang, Lei Zhang, Yezhou Yang, and Zicheng
  Liu.
\newblock Seed: Self-supervised distillation for visual representation.
\newblock In \emph{International Conference on Learning Representations},
  2021{\natexlab{c}}.

\bibitem[Futrzynski(2020)]{vit_fig}
Romain Futrzynski.
\newblock Getting meaning from text: self-attention step-by-step video.
\newblock \url{https://peltarion.com/blog/data-science/self-attention-video},
  2020.
\newblock Accessed: 2022-05-17.

\bibitem[Gidaris et~al.(2018)Gidaris, Singh, and
  Komodakis]{gidaris2018unsupervised}
Spyros Gidaris, Praveer Singh, and Nikos Komodakis.
\newblock Unsupervised representation learning by predicting image rotations.
\newblock \emph{International Conference on Learning Representations}, 2018.

\bibitem[Gou et~al.(2021)Gou, Yu, Maybank, and Tao]{gou2021knowledge}
Jianping Gou, Baosheng Yu, Stephen~J Maybank, and Dacheng Tao.
\newblock Knowledge distillation: A survey.
\newblock \emph{International Journal of Computer Vision}, 129\penalty0
  (6):\penalty0 1789--1819, 2021.

\bibitem[Graham et~al.(2021)Graham, El-Nouby, Touvron, Stock, Joulin,
  J{\'e}gou, and Douze]{graham2021levit}
Benjamin Graham, Alaaeldin El-Nouby, Hugo Touvron, Pierre Stock, Armand Joulin,
  Herv{\'e} J{\'e}gou, and Matthijs Douze.
\newblock Levit: a vision transformer in convnet's clothing for faster
  inference.
\newblock In \emph{Proceedings of the IEEE/CVF International Conference on
  Computer Vision}, pages 12259--12269, 2021.

\bibitem[Grill et~al.(2020)Grill, Strub, Altch{\'e}, Tallec, Richemond,
  Buchatskaya, Doersch, Avila~Pires, Guo, Gheshlaghi~Azar,
  et~al.]{grill2020bootstrap}
Jean-Bastien Grill, Florian Strub, Florent Altch{\'e}, Corentin Tallec, Pierre
  Richemond, Elena Buchatskaya, Carl Doersch, Bernardo Avila~Pires, Zhaohan
  Guo, Mohammad Gheshlaghi~Azar, et~al.
\newblock Bootstrap your own latent-a new approach to self-supervised learning.
\newblock \emph{Advances in Neural Information Processing Systems},
  33:\penalty0 21271--21284, 2020.

\bibitem[Hadsell et~al.(2006)Hadsell, Chopra, and
  LeCun]{hadsell2006dimensionality}
Raia Hadsell, Sumit Chopra, and Yann LeCun.
\newblock Dimensionality reduction by learning an invariant mapping.
\newblock In \emph{Proceedings of the IEEE Conference on Computer Vision and
  Pattern Recognition}, volume~2, pages 1735--1742. IEEE, 2006.

\bibitem[Hassani et~al.(2021)Hassani, Walton, Shah, Abuduweili, Li, and
  Shi]{hassani2021escaping}
Ali Hassani, Steven Walton, Nikhil Shah, Abulikemu Abuduweili, Jiachen Li, and
  Humphrey Shi.
\newblock Escaping the big data paradigm with compact transformers.
\newblock \emph{arXiv preprint arXiv:2104.05704}, 2021.

\bibitem[He et~al.(2020)He, Fan, Wu, Xie, and Girshick]{he2020momentum}
Kaiming He, Haoqi Fan, Yuxin Wu, Saining Xie, and Ross Girshick.
\newblock Momentum contrast for unsupervised visual representation learning.
\newblock In \emph{Proceedings of the IEEE/CVF conference on computer vision
  and pattern recognition}, pages 9729--9738, 2020.

\bibitem[He et~al.(2022{\natexlab{a}})He, Chen, Xie, Li, Doll{\'a}r, and
  Girshick]{he2021masked}
Kaiming He, Xinlei Chen, Saining Xie, Yanghao Li, Piotr Doll{\'a}r, and Ross
  Girshick.
\newblock Masked autoencoders are scalable vision learners.
\newblock \emph{Proceedings of the IEEE Conference on Computer Vision and
  Pattern Recognition}, 2022{\natexlab{a}}.

\bibitem[He et~al.(2022{\natexlab{b}})He, Sun, Yang, Bai, and
  Qi]{he2022knowledge}
Ruifei He, Shuyang Sun, Jihan Yang, Song Bai, and Xiaojuan Qi.
\newblock Knowledge distillation as efficient pre-training: Faster convergence,
  higher data-efficiency, and better transferability.
\newblock \emph{Proceedings of the IEEE Conference on Computer Vision and
  Pattern Recognition}, 2022{\natexlab{b}}.

\bibitem[Hinton et~al.(2014)Hinton, Vinyals, and Dean]{hinton2015distilling}
Geoffrey Hinton, Oriol Vinyals, and Jeff Dean.
\newblock Distilling the knowledge in a neural network.
\newblock \emph{Advances in Neural Information Processing Systems}, 2014.

\bibitem[Jia et~al.(2021)Jia, Han, Wang, Tang, Guo, Zhang, and
  Tao]{jia2021efficient}
Ding Jia, Kai Han, Yunhe Wang, Yehui Tang, Jianyuan Guo, Chao Zhang, and
  Dacheng Tao.
\newblock Efficient vision transformers via fine-grained manifold distillation.
\newblock \emph{arXiv preprint arXiv:2107.01378}, 2021.

\bibitem[Jiao et~al.(2020)Jiao, Yin, Shang, Jiang, Chen, Li, Wang, and
  Liu]{jiao2019tinybert}
Xiaoqi Jiao, Yichun Yin, Lifeng Shang, Xin Jiang, Xiao Chen, Linlin Li, Fang
  Wang, and Qun Liu.
\newblock Tinybert: Distilling bert for natural language understanding.
\newblock \emph{Empirical Methods in Natural Language Processing}, 2020.

\bibitem[Keys(1981)]{keys1981cubic}
Robert Keys.
\newblock Cubic convolution interpolation for digital image processing.
\newblock \emph{IEEE transactions on acoustics, speech, and signal processing},
  29\penalty0 (6):\penalty0 1153--1160, 1981.

\bibitem[Lee et~al.(2021)Lee, Lee, and Song]{lee2021vision}
Seung~Hoon Lee, Seunghyun Lee, and Byung~Cheol Song.
\newblock Vision transformer for small-size datasets.
\newblock \emph{arXiv preprint arXiv:2112.13492}, 2021.

\bibitem[Li et~al.(2022)Li, Yang, Zhang, Gao, Xiao, Dai, Yuan, and
  Gao]{li2022efficient}
Chunyuan Li, Jianwei Yang, Pengchuan Zhang, Mei Gao, Bin Xiao, Xiyang Dai,
  Lu~Yuan, and Jianfeng Gao.
\newblock Efficient self-supervised vision transformers for representation
  learning.
\newblock In \emph{International Conference on Learning Representations}, 2022.
\newblock URL \url{https://openreview.net/forum?id=fVu3o-YUGQK}.

\bibitem[Liu et~al.(2022)Liu, Liu, Li, and Liu]{liu2022meta}
Jihao Liu, Boxiao Liu, Hongsheng Li, and Yu~Liu.
\newblock Meta knowledge distillation.
\newblock \emph{arXiv preprint arXiv:2202.07940}, 2022.

\bibitem[Liu et~al.(2021)Liu, Lin, Cao, Hu, Wei, Zhang, Lin, and
  Guo]{liu2021swin}
Ze~Liu, Yutong Lin, Yue Cao, Han Hu, Yixuan Wei, Zheng Zhang, Stephen Lin, and
  Baining Guo.
\newblock Swin transformer: Hierarchical vision transformer using shifted
  windows.
\newblock In \emph{Proceedings of the IEEE/CVF International Conference on
  Computer Vision}, pages 10012--10022, 2021.

\bibitem[Loshchilov and Hutter(2019)]{loshchilov2017decoupled}
Ilya Loshchilov and Frank Hutter.
\newblock Decoupled weight decay regularization.
\newblock \emph{International Conference on Learning Representations}, 2019.

\bibitem[Mehta and Rastegari(2022)]{mehta2021mobilevit}
Sachin Mehta and Mohammad Rastegari.
\newblock Mobilevit: light-weight, general-purpose, and mobile-friendly vision
  transformer.
\newblock \emph{International Conference on Learning Representations}, 2022.

\bibitem[Navaneet et~al.(2021)Navaneet, Koohpayegani, Tejankar, and
  Pirsiavash]{navaneet2022simreg}
KL~Navaneet, Soroush~Abbasi Koohpayegani, Ajinkya Tejankar, and Hamed
  Pirsiavash.
\newblock Simreg: Regression as a simple yet effective tool for self-supervised
  knowledge distillation.
\newblock \emph{Proceedings of the British Machine Vision Conference}, 2021.

\bibitem[Noroozi et~al.(2018)Noroozi, Vinjimoor, Favaro, and
  Pirsiavash]{noroozi2018boosting}
Mehdi Noroozi, Ananth Vinjimoor, Paolo Favaro, and Hamed Pirsiavash.
\newblock Boosting self-supervised learning via knowledge transfer.
\newblock In \emph{Proceedings of the IEEE Conference on Computer Vision and
  Pattern Recognition}, pages 9359--9367, 2018.

\bibitem[Park et~al.(2019)Park, Kim, Lu, and Cho]{park2019relational}
Wonpyo Park, Dongju Kim, Yan Lu, and Minsu Cho.
\newblock Relational knowledge distillation.
\newblock In \emph{Proceedings of the IEEE/CVF Conference on Computer Vision
  and Pattern Recognition}, pages 3967--3976, 2019.

\bibitem[Pelosin et~al.(2022)Pelosin, Jha, Torsello, Raducanu, and van~de
  Weijer]{pelosin2022towards}
Francesco Pelosin, Saurav Jha, Andrea Torsello, Bogdan Raducanu, and Joost
  van~de Weijer.
\newblock Towards exemplar-free continual learning in vision transformers: an
  account of attention, functional and weight regularization.
\newblock \emph{Proceedings of the IEEE Conference on Computer Vision and
  Pattern Recognition}, 2022.

\bibitem[Peng et~al.(2019)Peng, Jin, Liu, Li, Wu, Liu, Zhou, and
  Zhang]{peng2019correlation}
Baoyun Peng, Xiao Jin, Jiaheng Liu, Dongsheng Li, Yichao Wu, Yu~Liu, Shunfeng
  Zhou, and Zhaoning Zhang.
\newblock Correlation congruence for knowledge distillation.
\newblock In \emph{Proceedings of the IEEE/CVF International Conference on
  Computer Vision}, pages 5007--5016, 2019.

\bibitem[Rebuffi et~al.(2017)Rebuffi, Kolesnikov, Sperl, and
  Lampert]{rebuffi2017icarl}
Sylvestre-Alvise Rebuffi, Alexander Kolesnikov, Georg Sperl, and Christoph~H
  Lampert.
\newblock icarl: Incremental classifier and representation learning.
\newblock In \emph{Proceedings of the IEEE Conference on Computer Vision and
  Pattern Recognition}, pages 2001--2010, 2017.

\bibitem[Romero et~al.(2014)Romero, Ballas, Kahou, Chassang, Gatta, and
  Bengio]{romero2014fitnets}
Adriana Romero, Nicolas Ballas, Samira~Ebrahimi Kahou, Antoine Chassang, Carlo
  Gatta, and Yoshua Bengio.
\newblock Fitnets: Hints for thin deep nets.
\newblock \emph{arXiv preprint arXiv:1412.6550}, 2014.

\bibitem[Russakovsky et~al.(2015)Russakovsky, Deng, Su, Krause, Satheesh, Ma,
  Huang, Karpathy, Khosla, Bernstein, et~al.]{russakovsky2015imagenet}
Olga Russakovsky, Jia Deng, Hao Su, Jonathan Krause, Sanjeev Satheesh, Sean Ma,
  Zhiheng Huang, Andrej Karpathy, Aditya Khosla, Michael Bernstein, et~al.
\newblock Imagenet large scale visual recognition challenge.
\newblock \emph{International Journal of Computer Vision}, 115\penalty0
  (3):\penalty0 211--252, 2015.

\bibitem[Strudel et~al.(2021)Strudel, Garcia, Laptev, and
  Schmid]{strudel2021segmenter}
Robin Strudel, Ricardo Garcia, Ivan Laptev, and Cordelia Schmid.
\newblock Segmenter: Transformer for semantic segmentation.
\newblock In \emph{Proceedings of the IEEE/CVF International Conference on
  Computer Vision}, pages 7262--7272, 2021.

\bibitem[Sun et~al.(2020)Sun, Yu, Song, Liu, Yang, and Zhou]{sun2020mobilebert}
Zhiqing Sun, Hongkun Yu, Xiaodan Song, Renjie Liu, Yiming Yang, and Denny Zhou.
\newblock Mobilebert: a compact task-agnostic bert for resource-limited
  devices.
\newblock \emph{Annual Meeting of the Association for Computational
  Linguistics}, 2020.

\bibitem[Sun et~al.(2021)Sun, Cao, Yang, and Kitani]{sun2021rethinking}
Zhiqing Sun, Shengcao Cao, Yiming Yang, and Kris~M Kitani.
\newblock Rethinking transformer-based set prediction for object detection.
\newblock In \emph{Proceedings of the IEEE/CVF International Conference on
  Computer Vision}, pages 3611--3620, 2021.

\bibitem[Tian et~al.(2020)Tian, Krishnan, and Isola]{tian2019contrastive}
Yonglong Tian, Dilip Krishnan, and Phillip Isola.
\newblock Contrastive representation distillation.
\newblock \emph{International Conference on Learning Representations}, 2020.

\bibitem[Touvron et~al.(2021{\natexlab{a}})Touvron, Cord, Douze, Massa,
  Sablayrolles, and Jegou]{deit}
Hugo Touvron, Matthieu Cord, Matthijs Douze, Francisco Massa, Alexandre
  Sablayrolles, and Herve Jegou.
\newblock Training data-efficient image transformers and distillation through
  attention.
\newblock In \emph{International Conference on Machine Learning}, volume 139,
  pages 10347--10357, July 2021{\natexlab{a}}.

\bibitem[Touvron et~al.(2021{\natexlab{b}})Touvron, Cord, Sablayrolles,
  Synnaeve, and J{\'e}gou]{touvron2021going}
Hugo Touvron, Matthieu Cord, Alexandre Sablayrolles, Gabriel Synnaeve, and
  Herv{\'e} J{\'e}gou.
\newblock Going deeper with image transformers.
\newblock In \emph{Proceedings of the IEEE/CVF International Conference on
  Computer Vision}, pages 32--42, 2021{\natexlab{b}}.

\bibitem[Touvron et~al.(2022)Touvron, Cord, and Jegou]{Touvron2022DeiTIR}
Hugo Touvron, Matthieu Cord, and Herve Jegou.
\newblock Deit iii: Revenge of the vit.
\newblock \emph{European Conference on Computer Vision}, 2022.

\bibitem[Tung and Mori(2019)]{tung2019similarity}
Frederick Tung and Greg Mori.
\newblock Similarity-preserving knowledge distillation.
\newblock In \emph{Proceedings of the IEEE/CVF International Conference on
  Computer Vision}, pages 1365--1374, 2019.

\bibitem[Van~den Oord et~al.(2018)Van~den Oord, Li, and
  Vinyals]{van2018representation}
Aaron Van~den Oord, Yazhe Li, and Oriol Vinyals.
\newblock Representation learning with contrastive predictive coding.
\newblock \emph{arXiv e-prints}, pages arXiv--1807, 2018.

\bibitem[Vaswani et~al.(2017)Vaswani, Shazeer, Parmar, Uszkoreit, Jones, Gomez,
  Kaiser, and Polosukhin]{vaswani2017attention}
Ashish Vaswani, Noam Shazeer, Niki Parmar, Jakob Uszkoreit, Llion Jones,
  Aidan~N Gomez, {\L}ukasz Kaiser, and Illia Polosukhin.
\newblock Attention is all you need.
\newblock \emph{Advances in neural information processing systems}, 30, 2017.

\bibitem[Wang et~al.(2021)Wang, Xie, Li, Fan, Song, Liang, Lu, Luo, and
  Shao]{wang2021pyramid}
Wenhai Wang, Enze Xie, Xiang Li, Deng-Ping Fan, Kaitao Song, Ding Liang, Tong
  Lu, Ping Luo, and Ling Shao.
\newblock Pyramid vision transformer: A versatile backbone for dense prediction
  without convolutions.
\newblock In \emph{Proceedings of the IEEE/CVF International Conference on
  Computer Vision}, pages 568--578, 2021.

\bibitem[Wang et~al.(2020)Wang, Wei, Dong, Bao, Yang, and Zhou]{wang2020minilm}
Wenhui Wang, Furu Wei, Li~Dong, Hangbo Bao, Nan Yang, and Ming Zhou.
\newblock Minilm: Deep self-attention distillation for task-agnostic
  compression of pre-trained transformers.
\newblock \emph{Advances in Neural Information Processing Systems},
  33:\penalty0 5776--5788, 2020.

\bibitem[Wu et~al.(2021)Wu, Xiao, Codella, Liu, Dai, Yuan, and
  Zhang]{wu2021cvt}
Haiping Wu, Bin Xiao, Noel Codella, Mengchen Liu, Xiyang Dai, Lu~Yuan, and Lei
  Zhang.
\newblock Cvt: Introducing convolutions to vision transformers.
\newblock In \emph{Proceedings of the IEEE/CVF International Conference on
  Computer Vision}, pages 22--31, 2021.

\bibitem[Wu et~al.(2018)Wu, Xiong, Yu, and Lin]{wu2018unsupervised}
Zhirong Wu, Yuanjun Xiong, Stella~X Yu, and Dahua Lin.
\newblock Unsupervised feature learning via non-parametric instance
  discrimination.
\newblock In \emph{Proceedings of the IEEE conference on computer vision and
  pattern recognition}, pages 3733--3742, 2018.

\bibitem[Xie et~al.(2021{\natexlab{a}})Xie, Wang, Yu, Anandkumar, Alvarez, and
  Luo]{xie2021segformer}
Enze Xie, Wenhai Wang, Zhiding Yu, Anima Anandkumar, Jose~M Alvarez, and Ping
  Luo.
\newblock Segformer: Simple and efficient design for semantic segmentation with
  transformers.
\newblock \emph{Advances in Neural Information Processing Systems}, 34,
  2021{\natexlab{a}}.

\bibitem[Xie et~al.(2021{\natexlab{b}})Xie, Lin, Yao, Zhang, Dai, Cao, and
  Hu]{xie2021self}
Zhenda Xie, Yutong Lin, Zhuliang Yao, Zheng Zhang, Qi~Dai, Yue Cao, and Han Hu.
\newblock Self-supervised learning with swin transformers.
\newblock \emph{arXiv preprint arXiv:2105.04553}, 2021{\natexlab{b}}.

\bibitem[Ye and Bors(2021)]{ye2021lifelong}
Fei Ye and Adrian Bors.
\newblock Lifelong teacher-student network learning.
\newblock \emph{{IEEE} Transactions on Pattern Analysis and Machine
  Intelligence}, 2021.

\bibitem[Yu et~al.(2019)Yu, Yazici, Liu, Weijer, Cheng, and
  Ramisa]{yu2019learning}
Lu~Yu, Vacit~Oguz Yazici, Xialei Liu, Joost van~de Weijer, Yongmei Cheng, and
  Arnau Ramisa.
\newblock Learning metrics from teachers: Compact networks for image embedding.
\newblock In \emph{Proceedings of the IEEE/CVF Conference on Computer Vision
  and Pattern Recognition}, pages 2907--2916, 2019.

\bibitem[Yu et~al.(2022)Yu, Chen, Shen, Yuan, Tan, Yang, Liu, and
  Wang]{yu2022unified}
Shixing Yu, Tianlong Chen, Jiayi Shen, Huan Yuan, Jianchao Tan, Sen Yang,
  Ji~Liu, and Zhangyang Wang.
\newblock Unified visual transformer compression.
\newblock \emph{International Conference on Learning Representations}, 2022.

\bibitem[Yuan et~al.(2020)Yuan, Tay, Li, Wang, and Feng]{yuan2020revisiting}
Li~Yuan, Francis~EH Tay, Guilin Li, Tao Wang, and Jiashi Feng.
\newblock Revisiting knowledge distillation via label smoothing regularization.
\newblock In \emph{Proceedings of the IEEE/CVF Conference on Computer Vision
  and Pattern Recognition}, pages 3903--3911, 2020.

\bibitem[Yuan et~al.(2021)Yuan, Chen, Wang, Yu, Shi, Jiang, Tay, Feng, and
  Yan]{yuan2021tokens}
Li~Yuan, Yunpeng Chen, Tao Wang, Weihao Yu, Yujun Shi, Zi-Hang Jiang,
  Francis~EH Tay, Jiashi Feng, and Shuicheng Yan.
\newblock Tokens-to-token vit: Training vision transformers from scratch on
  imagenet.
\newblock In \emph{Proceedings of the IEEE/CVF International Conference on
  Computer Vision}, pages 558--567, 2021.

\bibitem[Zagoruyko and Komodakis(2017)]{zagoruyko2016paying}
Sergey Zagoruyko and Nikos Komodakis.
\newblock Paying more attention to attention: Improving the performance of
  convolutional neural networks via attention transfer.
\newblock \emph{International Conference on Learning Representations}, 2017.

\bibitem[Zhao et~al.(2022)Zhao, Cui, Song, Qiu, and Liang]{zhao2022decoupled}
Borui Zhao, Quan Cui, Renjie Song, Yiyu Qiu, and Jiajun Liang.
\newblock Decoupled knowledge distillation.
\newblock \emph{arXiv preprint arXiv:2203.08679}, 2022.

\bibitem[Zheng et~al.(2021)Zheng, Lu, Zhao, Zhu, Luo, Wang, Fu, Feng, Xiang,
  Torr, et~al.]{zheng2021rethinking}
Sixiao Zheng, Jiachen Lu, Hengshuang Zhao, Xiatian Zhu, Zekun Luo, Yabiao Wang,
  Yanwei Fu, Jianfeng Feng, Tao Xiang, Philip~HS Torr, et~al.
\newblock Rethinking semantic segmentation from a sequence-to-sequence
  perspective with transformers.
\newblock In \emph{Proceedings of the IEEE/CVF conference on computer vision
  and pattern recognition}, pages 6881--6890, 2021.

\bibitem[Zhou et~al.(2022{\natexlab{a}})Zhou, Wei, Wang, Shen, Xie, Yuille, and
  Kong]{zhou2021ibot}
Jinghao Zhou, Chen Wei, Huiyu Wang, Wei Shen, Cihang Xie, Alan Yuille, and Tao
  Kong.
\newblock ibot: Image bert pre-training with online tokenizer.
\newblock \emph{International Conference on Learning Representations (ICLR)},
  2022{\natexlab{a}}.

\bibitem[Zhou et~al.(2022{\natexlab{b}})Zhou, Zhou, Si, Yu, Ng, and
  Yan]{zhou2022mugs}
Pan Zhou, Yichen Zhou, Chenyang Si, Weihao Yu, Teck~Khim Ng, and Shuicheng Yan.
\newblock Mugs: A multi-granular self-supervised learning framework.
\newblock \emph{arXiv preprint arXiv:2203.14415}, 2022{\natexlab{b}}.

\bibitem[Zhu et~al.(2021)Zhu, Su, Lu, Li, Wang, and Dai]{zhu2020deformable}
Xizhou Zhu, Weijie Su, Lewei Lu, Bin Li, Xiaogang Wang, and Jifeng Dai.
\newblock Deformable detr: Deformable transformers for end-to-end object
  detection.
\newblock \emph{International Conference on Learning Representations}, 2021.

\end{thebibliography}



\section*{Supplementary material}

# Supplementary Materials:
# Attention Distillation: self-supervised vision transformer students need more guidance

Kai Wang
kwang@cvc.uab.es

Fei Yang (*corresponding author)
fyang@cvc.uab.es

Joost van de Weijer
joost@cvc.uab.es

Computer Vision Center
Universitat Autònoma de Barcelona
Barcelona, Spain

## 1 Table 2 visualization

A graphical visualization for Table 2 in the main paper is shown in Fig. 6, where you can easily observe the performance difference between each teacher and student pair. We can also observe that the performance drop from linear probing to $k$-NN is effectively reduced by AttnDistill.

## 2 Networks configurations

In this paper, our networks configurations mainly refer to the ViT designs in DINO [1], DEiT [9], Mugs [12] and iBOT [11], where the number of heads $H$ and position embedding (PE) strategy (learnable PE and ViT-S with 6 head) are different from MoCo v3 [3] (fixed sin-cos PE and ViT-S with 12 head). The detailed networks configurations are shown in Table 7.

## 3 Additional Implementations

### 3.1 Pre-Training recipe

A clear pre-training recipe is shown in Table 8. We mainly refer to the training recipe from MAE [5].

### 3.2 More details for evaluations

***k*-NN, Linear Probing and finetuning on ImageNet-1K.** To evaluate the quality of pre-trained features, we either use a k-nearest neighbor (k-NN) classifier or a linear classifier

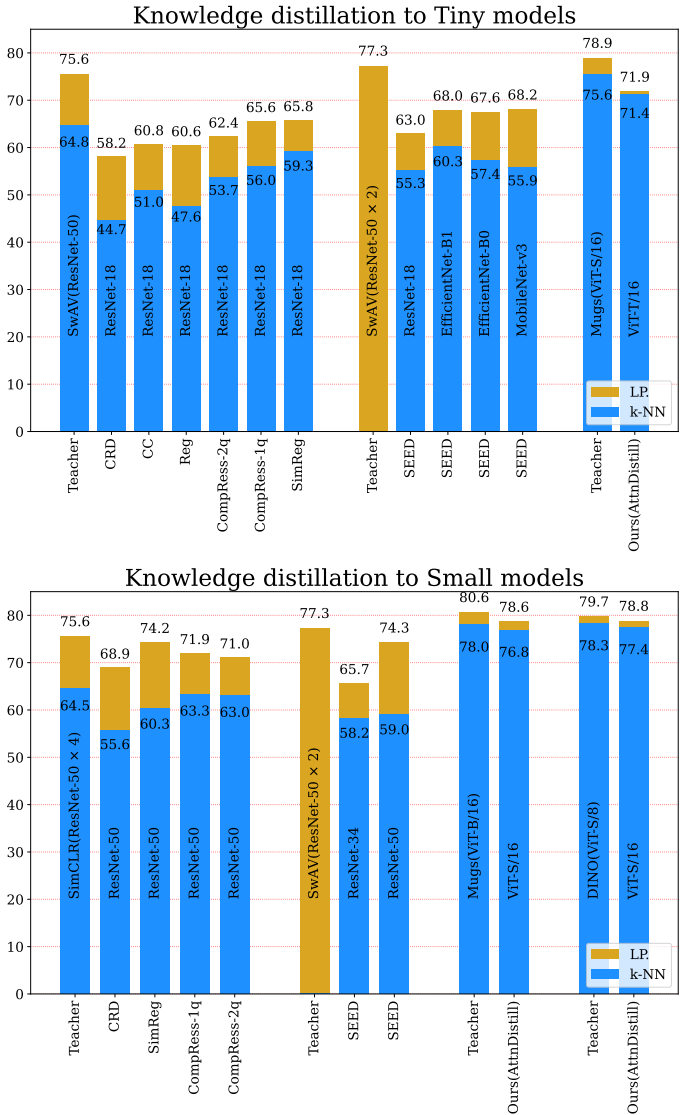

Figure 6: Visualization of Table 2 results, providing both the linear probing and *k*-NN accuracies for multiple methods. Note that AttnDistill significantly reduces the gap with the teacher model.

| | PE | model | layers | dim | heads | patch size | #tokens | #params |
|---|---|---|---|---|---|---|---|---|
| | | Networks configurations for experiments on **ImageNet-1K** | | | | | | |
| Teacher | learnable | Mugs (ViT-S/16) | 12 | 384 | 6 | 16 | 197 | 22M |
| | | DINO (ViT-S/8) | 12 | 384 | 6 | 8 | 785 | 22M |
| | | Mugs (ViT-B/16) | 12 | 768 | 12 | 16 | 197 | 85M |
| Student | learnable | AttnDistill (ViT-T/16) | 12 | 192 | 3 | 16 | 197 | 5.7M |
| | | AttnDistill (ViT-S/16) | 12 | 384 | 6 | 16 | 197 | 22M |
| | | Networks configurations for experiments on **ImageNet-Subset** | | | | | | |
| Teacher | sin-cos | MAE (ViT-S/16) | 12 | 384 | 6 | 16 | 197 | 22M |
| Student | sin-cos | AttnDistill (ViT-T) | 12 | 192 | 6 | 16 | 197 | 5.7M |
| | | AttnDistill (ViT-T) | 12 | 192 | 3 | 16 | 197 | 5.7M |
| | | AttnDistill (ViT-T) | 12 | 192 | 3 | 32 | 65 | 5.7M |
| | | AttnDistill (ViT-T) | 8 | 192 | 3 | 16 | 197 | 3.8M |
| | | AttnDistill (ViT-T) | 8 | 192 | 3 | 32 | 65 | 3.8M |

Table 7: **Networks configuration.** "layers" is the number of Transformer blocks, "dim" is channel dimension and "heads" is the number of heads in multi-head attention. "# tokens" is the length of the token sequence, "# params" is the total number of parameters. "PE" is the position embedding strategy. We consider $224 \times 224$ resolution inputs.

| config | value |
|---|---|
| optimizer | AdamW [8] |
| base learning rate | 1.5e-4 |
| weight decay | 0.05 |
| optimizer momentum | $\beta_1, \beta_2 = 0.9, 0.95$ [2] |
| batch size | 4096 |
| learning rate schedule | cosine decay [7] |
| warmup epochs | 40 |
| training epochs | 500 (ViT-T/16 ImageNet-1K) |
| | 800 (ViT-S/16 ImageNet-1K) |
| | 3200 (ViT-T/16 ImageNet-Subset) |
| augmentation | RandomResizedCrop |

Table 8: Pre-Training settings for ViTs distillation on **ImageNet-1K** and **ImageNet-Subset**.

| config | value |
|---|---|
| optimizer | LARS [10] |
| base learning rate | 0.1 |
| weight decay | 0 |
| optimizer momentum | 0.9 |
| batch size | 16384 |
| learning rate schedule | cosine decay |
| warmup epochs | 10 |
| training epochs | 90 |
| augmentation | RandomResizedCrop |

Table 9: **Linear probing setting** on **ImageNet-Subset** for self-supervised knowledge distillation with a MAE(ViT-S/16) teacher.

on the frozen representation. We follow the evaluation protocols in DINO [1], iBOT [11], Mugs [12]. For k-NN evaluation, we sweep over different numbers of nearest neighbors. For linear probing and finetuning evaluations, we sweep over different learning rates.

**Semi-supervised learning on ImageNet-1K.** In this setting, first, models are trained self-supervised on all *ImageNet-1K* data. Next, labels for a small fraction of data (1% or 10%) are used to perform fine-tuning, linear probing or *k*-NN classification. This is also an *extension* to the semi-supervised learning evaluations in DINO, iBOT and Mugs papers. For k-NN evaluation, we sweep over different numbers of nearest neighbors. For linear evaluation, we sweep over different learning rates. For fine-tuning evaluation, we fine-tune the pre-trained backbone for 1000 epochs with learning rate set to 5e-6.

**Transfer learning.** We pretrain the model on ImageNet-1K, and then fine-tune the pre-trained backbone on various datasets with the same protocols and optimization settings as in DINO, iBOT, and Mugs. For both ViT-T and ViT-S, we use AdamW optimizer with a minibatch of 1024, we fine-tune the pretrained model 1000 epochs by sweeping the learning rates. The weight decay is fixed to be 0.05.

**Linear probing evaluation on ImageNet-Subset** Since on ImageNet-Subset we consider the MAE as the teacher, thus in linear probing, we also follow the evaluation protocols from MAE [5] as shown in Table 9.

# 4 MAE pretraining on ImageNet-Subset

The linear probing curves of MAE [5] pre-training on ImageNet-Subset is shown in Fig. 7 for further references (from 1200 epochs to 3200 epochs). We can observe that both ViT-S(12-layers, 6-heads, 16-patches, 384-dim) and ViT-T(12-layers, 6-heads, 16-patches, 192-dim) both are saturated at 3200 epochs.

# 5 Additional ablation studies.

To prove the generalizability of AttnDistill, we perform the ablation study on ImageNet-Subset with a fixed MAE(ViT-S/16) teacher and vary the architecture of the student model. Apart from the ablation studies shown in Fig.5 in the main paper, here we extend the ablation studies and also display all the numbers included in Fig.5. As can be seen from Table 10, our ablation study can be roughly divided into six parts. The beginning three parts (a)-(c) are corresponding to the Fig.5(a)-(c). Except that, we further show our ablation study on the following three aspects:

- The design of the linear mapping $\mathcal{P}$: In Table 10-(d), we vary the number of $\mathcal{P}$ layers in $\{1,2,4,8\}$ and evaluate the output features from each layer. As can be seen that, the output feature before $\mathcal{P}$ (indicated by a 0 in column *Evaluation P layer*) is always a good choice for all considered layer-numbers variations. Also, for the number of layers 4 is a better choice.

- The hyperparameter of AttnDistill: In Table 10-(e) and Fig. 8, we ablate the $\lambda$ and $T$. The optimal choice is $\lambda = 0.1, T = 10.0$.

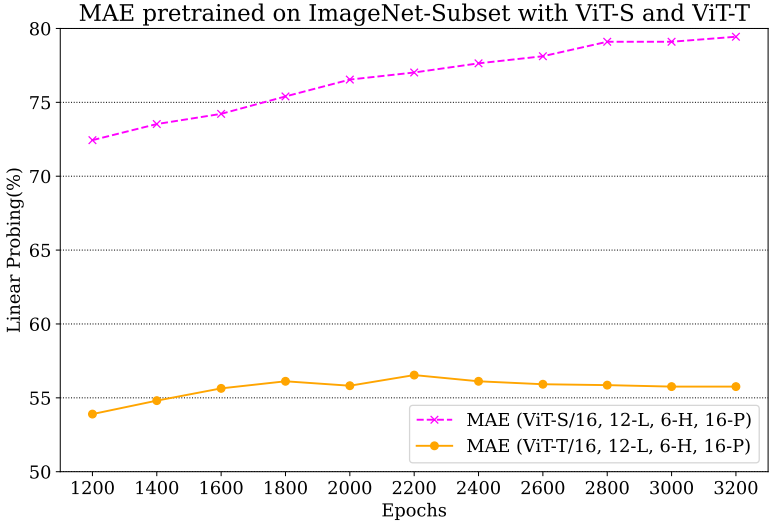

Figure 7: MAE pretrained on ImageNet-Subset with ViT-S and ViT-T models.

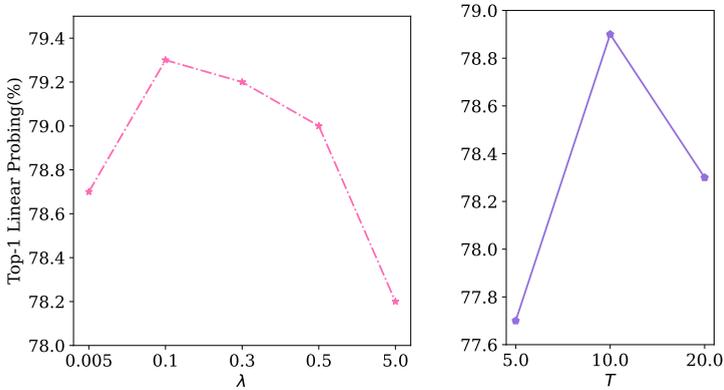

Figure 8: Ablation study over $T$ and $\lambda$.

- Other ablation studies: here we fix the design of the student model and vary the strategy to compute the attention distillation loss. We imitate the ATS [4] to compute the token scores with the norm of *Value* vectors in the MSA module of ViTs. We also replace the KL divergence in $\mathcal{L}_a$ with MSE loss. Neither of them could work better than our solution.

Dataset: ImageNet-Subset

Teacher: *ViT-S* (12-Layer, 6-Head, 16-Patch, 3200 epo., 384 dim); Top-1 LP.: 79.4; Top-5 LP.: 93.6

Student: *ViT-T* (12-Layer, 6-Head, 16-Patch, 3200 epo., 192 dim); Top-1 LP.: 63.7; Top-5 LP.: 86.2

| | Study on | Layers | Heads | Patch size | PA | Patch token align | CompRess KD | P layers | Attn Layer | AttnDistill | Attn Aggr | Attn Interpolate | Attn λ | Attn T | Evaluation P layer | Top-1 LP. | Top-5 LP. |
|---|---|---|---|---|---|---|---|---|---|---|---|---|---|---|---|---|---|
| (a) | Heads | 12 | 3 | 16 | ✓ | ✗ | ✗ | 4 | - | ✗ | ✗ | ✗ | - | - | 0 | 77.8 | 93.6 |
| | | 12 | 3 | 16 | ✓ | ✗ | ✗ | 4 | last | ✓ | ✓ | ✗ | 0.1 | 10.0 | 0 | 78.9 | 93.8 |
| | | 12 | 6 | 16 | ✓ | ✗ | ✗ | 4 | - | ✗ | ✗ | ✗ | - | - | 0 | 77.8 | 93.3 |
| | | 12 | 6 | 16 | ✓ | ✗ | ✗ | 4 | last | ✓ | ✓ | ✗ | 0.1 | 10.0 | 0 | 78.9 | 93.9 |
| | | 12 | 6 | 16 | ✓ | ✗ | ✗ | 4 | last | ✓ | ✗ | ✗ | 0.1 | - | 0 | **79.3** | **94.1** |
| | Patch size | 12 | 6 | 28 | ✓ | ✗ | ✗ | 4 | - | ✗ | ✗ | ✗ | - | - | 0 | 73.3 | 91.1 |
| | | 12 | 6 | 28 | ✓ | ✗ | ✗ | 4 | last | ✓ | ✗ | ✗ | 0.1 | - | 0 | 77.3 | 93.1 |
| | | 12 | 6 | 32 | ✓ | ✗ | ✗ | 4 | - | ✗ | ✗ | ✗ | - | - | 0 | 73.1 | 91.0 |
| | | 12 | 6 | 32 | ✓ | ✗ | ✗ | 4 | last | ✓ | ✗ | ✗ | 0.1 | - | 0 | 77.2 | 92.8 |
| | Layers | 8 | 6 | 16 | ✓ | ✗ | ✗ | 4 | - | ✗ | ✗ | ✗ | - | - | 0 | 77.7 | 92.8 |
| | | 8 | 6 | 16 | ✓ | ✗ | ✗ | 4 | last | ✓ | ✗ | ✗ | 0.1 | - | 0 | 79.1 | 94.0 |
| | | 8 | 3 | 32 | ✓ | ✗ | ✗ | 4 | - | ✗ | ✗ | ✗ | - | - | 0 | 68.6 | 88.8 |
| | | 8 | 3 | 32 | ✓ | ✗ | ✗ | 4 | last | ✓ | ✓ | ✓ | 0.1 | 10.0 | 0 | 73.8 | 91.7 |
| (b) | PA | 8 | 3 | 32 | ✓ | ✗ | ✗ | 4 | - | ✗ | ✗ | ✗ | - | - | 0 | 68.6 | 88.8 |
| | AttnDistill | 8 | 3 | 32 | ✓ | ✗ | ✗ | 4 | last | ✓ | ✓ | ✓ | 0.1 | 10.0 | 0 | 73.8 | 91.7 |
| | MAX | 8 | 3 | 32 | ✓ | ✗ | ✗ | 4 | last | ✓ | ✓ | ✓ | 0.1 | 10.0 | 0 | 73.5 | 91.6 |
| | MEAN | 8 | 3 | 32 | ✓ | ✗ | ✗ | 4 | last | ✓ | ✓ | ✓ | 0.1 | 10.0 | 0 | 70.6 | 90.0 |
| | MIN | 8 | 3 | 32 | ✓ | ✗ | ✗ | 4 | last | ✓ | ✓ | ✓ | 0.1 | 10.0 | 0 | 71.9 | 90.7 |
| (c) | KD | 12 | 6 | 16 | ✓ | ✗ | ✓ | 4 | - | ✗ | ✗ | ✗ | - | - | 0 | 75.0 | 92.5 |
| | | 12 | 6 | 16 | ✗ | ✗ | ✓ | 4 | - | ✗ | ✗ | ✗ | - | - | 0 | 71.7 | 90.7 |
| | Patch | 12 | 6 | 16 | ✗ | ✓ | ✗ | 4 | - | ✗ | ✗ | ✗ | - | - | 0 | 73.6 | 91.4 |
| | | 12 | 6 | 16 | ✓ | ✓ | ✗ | 4 | - | ✗ | ✗ | ✗ | - | - | 0 | 76.9 | 93.4 |
| | Attn Layer | 12 | 6 | 16 | ✓ | ✗ | ✗ | 4 | all | ✓ | ✗ | ✗ | 0.1 | - | 0 | 78.7 | 93.7 |
| | | 12 | 6 | 16 | ✓ | ✗ | ✗ | 4 | last | ✓ | ✗ | ✗ | 0.1 | - | 0 | **79.3** | **94.1** |
| (d) | MLP layers and Eval layer | 12 | 6 | 16 | ✓ | ✗ | ✗ | 8 | - | ✗ | ✗ | ✗ | - | - | 0 | 76.2 | 92.4 |
| | | 12 | 6 | 16 | ✓ | ✗ | ✗ | 1 | - | ✗ | ✗ | ✗ | - | - | 0 | 76.3 | 92.5 |
| | | 12 | 6 | 16 | ✓ | ✗ | ✗ | 1 | - | ✗ | ✗ | ✗ | - | - | 1 | 76.2 | 92.4 |
| | | 12 | 6 | 16 | ✓ | ✗ | ✗ | 2 | - | ✗ | ✗ | ✗ | - | - | 0 | 76.4 | 93.2 |
| | | 12 | 6 | 16 | ✓ | ✗ | ✗ | 2 | - | ✗ | ✗ | ✗ | - | - | 1 | 76.4 | 93.1 |
| | | 12 | 6 | 16 | ✓ | ✗ | ✗ | 2 | - | ✗ | ✗ | ✗ | - | - | 2 | 76.5 | 93.1 |
| | | 12 | 6 | 16 | ✓ | ✗ | ✗ | 4 | - | ✗ | ✗ | ✗ | - | - | 0 | 77.8 | 93.3 |
| | | 12 | 6 | 16 | ✓ | ✗ | ✗ | 4 | - | ✗ | ✗ | ✗ | - | - | 1 | 77.8 | 93.3 |
| | | 12 | 6 | 16 | ✓ | ✗ | ✗ | 4 | - | ✗ | ✗ | ✗ | - | - | 2 | 77.2 | 93.2 |
| | | 12 | 6 | 16 | ✓ | ✗ | ✗ | 4 | - | ✗ | ✗ | ✗ | - | - | 3 | 76.6 | 92.9 |
| | | 12 | 6 | 16 | ✓ | ✗ | ✗ | 4 | - | ✗ | ✗ | ✗ | - | - | 4 | 76.2 | 92.8 |
| (e) | Attn λ | 12 | 6 | 16 | ✓ | ✗ | ✗ | 4 | last | ✓ | ✗ | ✗ | 0.005 | - | 0 | 78.7 | 93.7 |
| | | 12 | 6 | 16 | ✓ | ✗ | ✗ | 4 | last | ✓ | ✗ | ✗ | 0.1 | - | 0 | **79.3** | **94.1** |
| | | 12 | 6 | 16 | ✓ | ✗ | ✗ | 4 | last | ✓ | ✗ | ✗ | 0.3 | - | 0 | 79.2 | 94.0 |
| | | 12 | 6 | 16 | ✓ | ✗ | ✗ | 4 | last | ✓ | ✗ | ✗ | 0.5 | - | 0 | 79.0 | 93.8 |
| | | 12 | 6 | 16 | ✓ | ✗ | ✗ | 4 | last | ✓ | ✗ | ✗ | 5.0 | - | 0 | 78.2 | 93.6 |
| | Attn T | 12 | 3 | 16 | ✓ | ✗ | ✗ | 4 | last | ✓ | ✓ | ✗ | 0.1 | 10.0 | 0 | 78.9 | 93.8 |
| | | 12 | 3 | 16 | ✓ | ✗ | ✗ | 4 | last | ✓ | ✓ | ✗ | 0.1 | 5.0 | 0 | 77.7 | 93.6 |
| | | 12 | 3 | 16 | ✓ | ✗ | ✗ | 4 | last | ✓ | ✓ | ✗ | 0.1 | 20.0 | 0 | 78.3 | 93.7 |
| (f) | PA | 8 | 3 | 32 | ✓ | ✗ | ✗ | 4 | - | ✗ | ✗ | ✗ | - | - | 0 | 68.6 | 88.8 |
| | Weight by \|VALUE\| [4] | 8 | 3 | 32 | ✓ | ✗ | ✗ | 4 | last | ✓ | ✓ | ✓ | 0.1 | 10.0 | 0 | 70.6 | 90.5 |
| | MSE loss [6] | 8 | 3 | 32 | ✓ | ✗ | ✗ | 4 | last | ✓ | ✓ | ✓ | 0.1 | 10.0 | 0 | 72.9 | 90.9 |
| | AttnDistill | 8 | 3 | 32 | ✓ | ✗ | ✗ | 4 | last | ✓ | ✓ | ✓ | 0.1 | 10.0 | 0 | 73.8 | 91.7 |

Table 10: Full table for our ablation studies.

# 6    More visualization of attention maps

Except the Fig.4 in the main paper, here in Fig. 9 and Fig. 10 we also show more visualization of the attention maps obtained from various knowledge distillation methods (MAE-ViT-S →ViT-T) on ImageNet-Subset. More attention visualizations with nearly 1000 images (≈ 10 images per class) are in our supplementary file named *"attn_vis.zip"* with the same layouts.

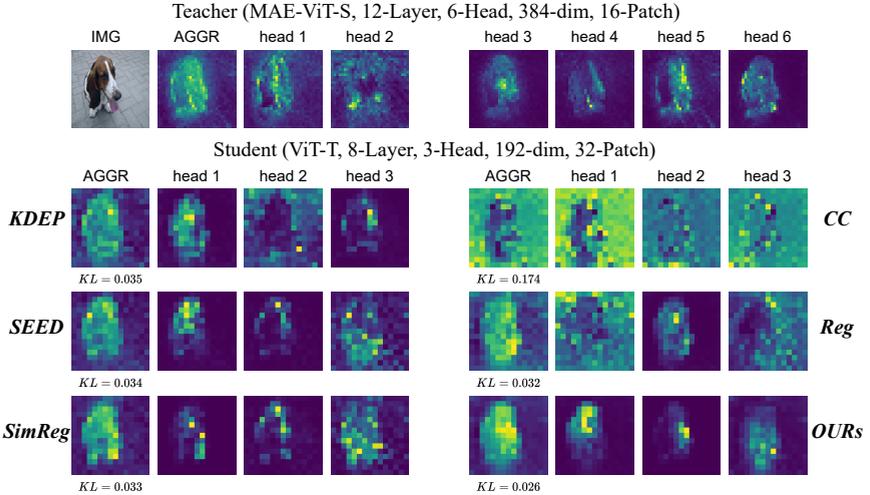

Figure 9: Comparison on attention maps for *"n02088238_194.jpg"*. For the teacher model ViT-S trained with MAE, we show the original image (IMG), the aggregated attention map (AGGR) with our AttnDistill and the attention maps for each head. For the student model ViT-T distilled from the teacher model, we show the aggregated attention map and each head attention map for each method. The KL distances to the teacher aggregated attention maps are shown under each method.

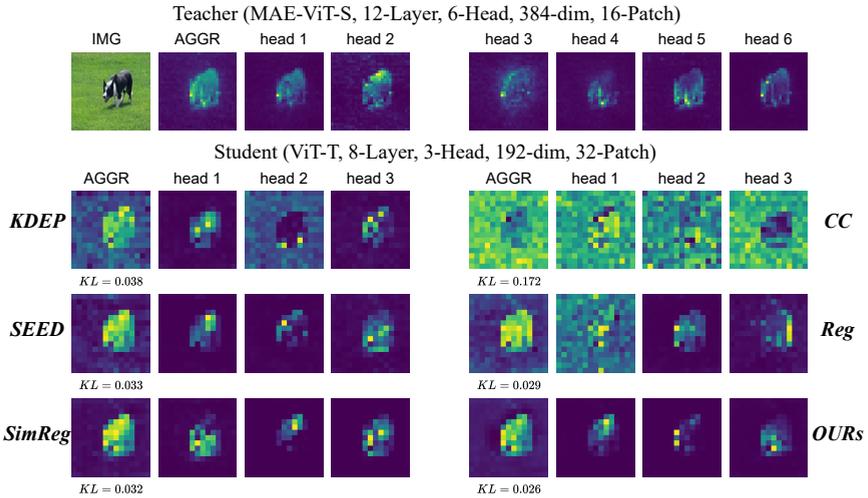

Figure 10: Comparison on attention maps for *"n02106166_45.jpg"*. For the teacher model ViT-S trained with MAE, we show the original image (IMG), the aggregated attention map (AGGR) with our AttnDistill and the attention maps for each head. For the student model ViT-T distilled from the teacher model, we show the aggregated attention map and each head attention map for each method. The KL distances to the teacher aggregated attention maps are shown under each method.

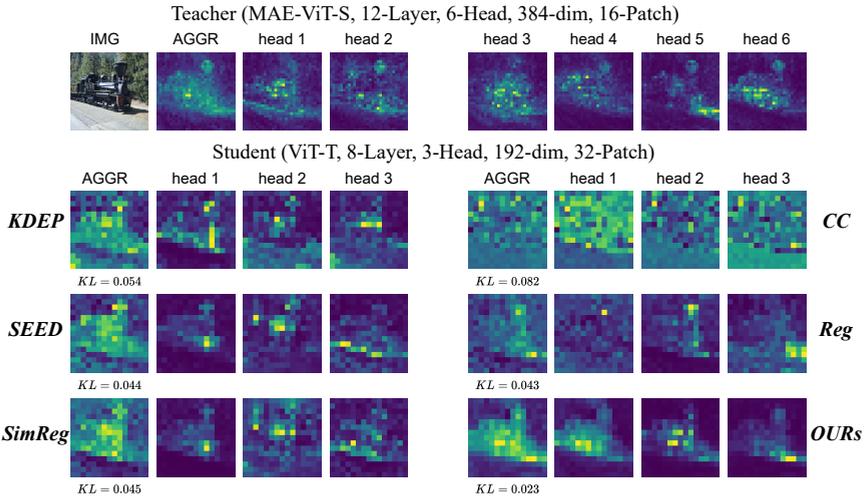

Figure 11: Comparison on attention maps *"n04310018_78.jpg"*. For the teacher model ViT-S trained with MAE, we show the original image (IMG), the aggregated attention map (AGGR) with our AttnDistill and the attention maps for each head. For the student model ViT-T distilled from the teacher model, we show the aggregated attention map and each head attention map for each method. The KL distances to the teacher aggregated attention maps are shown under each method.

# References

[1] Mathilde Caron, Hugo Touvron, Ishan Misra, Hervé Jégou, Julien Mairal, Piotr Bojanowski, and Armand Joulin. Emerging properties in self-supervised vision transformers. In *Proceedings of the IEEE/CVF International Conference on Computer Vision*, pages 9650–9660, 2021.

[2] Mark Chen, Alec Radford, Rewon Child, Jeffrey Wu, Heewoo Jun, David Luan, and Ilya Sutskever. Generative pretraining from pixels. In *International Conference on Machine Learning*, pages 1691–1703. PMLR, 2020.

[3] Xinlei Chen, Saining Xie, and Kaiming He. An empirical study of training self-supervised vision transformers. In *Proceedings of the IEEE/CVF International Conference on Computer Vision*, pages 9640–9649, 2021.

[4] Mohsen Fayyaz, Soroush Abbasi Kouhpayegani, Farnoush Rezaei Jafari, Eric Sommerlade, Hamid Reza Vaezi Joze, Hamed Pirsiavash, and Juergen Gall. Ats: Adaptive token sampling for efficient vision transformers. *arXiv preprint arXiv:2111.15667*, 2021.

[5] Kaiming He, Xinlei Chen, Saining Xie, Yanghao Li, Piotr Dollár, and Ross Girshick. Masked autoencoders are scalable vision learners. *Proceedings of the IEEE Conference on Computer Vision and Pattern Recognition*, 2022.

[6] Taehyeon Kim, Jaehoon Oh, NakYil Kim, Sangwook Cho, and Se-Young Yun. Comparing kullback-leibler divergence and mean squared error loss in knowledge distillation. *Proceedings of the International Joint Conference on Artificial Intelligence*, 2021.

[7] Ilya Loshchilov and Frank Hutter. Sgdr: Stochastic gradient descent with warm restarts. *International Conference on Learning Representations*, 2017.

[8] Ilya Loshchilov and Frank Hutter. Decoupled weight decay regularization. *International Conference on Learning Representations*, 2019.

[9] Hugo Touvron, Matthieu Cord, Matthijs Douze, Francisco Massa, Alexandre Sablayrolles, and Herve Jegou. Training data-efficient image transformers and distillation through attention. In *International Conference on Machine Learning*, volume 139, pages 10347–10357, July 2021.

[10] Yang You, Igor Gitman, and Boris Ginsburg. Large batch training of convolutional networks. *arXiv preprint arXiv:1708.03888*, 2017.

[11] Jinghao Zhou, Chen Wei, Huiyu Wang, Wei Shen, Cihang Xie, Alan Yuille, and Tao Kong. ibot: Image bert pre-training with online tokenizer. *International Conference on Learning Representations (ICLR)*, 2022.

[12] Pan Zhou, Yichen Zhou, Chenyang Si, Weihao Yu, Teck Khim Ng, and Shuicheng Yan. Mugs: A multi-granular self-supervised learning framework. *arXiv preprint arXiv:2203.14415*, 2022.



\end{document}